\documentclass[twocolumn]{journal}
\usepackage{arxiv}
\usepackage[utf8]{inputenc} 
\usepackage[T1]{fontenc}    
\usepackage{hyperref}       
\usepackage{url}            
\usepackage{booktabs}       
\usepackage{amsfonts}       
\usepackage{nicefrac}       
\usepackage{microtype}      
\usepackage{lipsum}
\usepackage{graphicx}       
\usepackage{tabularx}       
\usepackage{float}
\usepackage{xargs}
\usepackage{amsmath}
\usepackage{amsfonts}
\usepackage{amsthm}
\usepackage{acronym} 
\usepackage{babel}
\usepackage{color}          
\usepackage{multirow}       
\usepackage{numprint}
\usepackage{stackengine}    
\usepackage{siunitx}
\usepackage{caption}        
\usepackage{subcaption}     
\usepackage{placeins}
\usepackage{dsfont}

\DeclareSIUnit{\euro}{\text{€}}
\DeclareSIUnit{\euroMW}{\text{€/MW}}
\DeclareSIUnit{\euroMWh}{(\text{€/MW)/h}}
\DeclareSIUnit{\euroMWhsqr}{\text{(€/MW)/h)\textsuperscript{2}}}

\usepackage{amsmath}
\DeclareMathOperator*{\argmax}{arg\,max}

\npdecimalsign{.} 

\graphicspath{ {./images/} }

\title{ML-Based Bidding Price Prediction for Pay-As-Bid Ancillary Services Markets: A Use Case in the German Control Reserve Market}

\author{
 Vincent Bezold\textsuperscript{1, *}\\
  Fraunhofer Institute for Manufacturing Engineering and Automation IPA,\\
  Nobelstraße 12, Stuttgart, 70569, Germany\\
  ORCID: 0009-0006-4430-4710 \\
  \texttt{vincent.bezold@ipa.fraunhofer.de} \\
  \And
 Lukas Baur\textsuperscript{1}\\
  Fraunhofer Institute for Manufacturing Engineering and Automation IPA,\\
  Nobelstraße 12, Stuttgart, 70569, Germany\\
  ORCID: 0000-0002-6265-9877 \\
  \texttt{lukas.baur@ipa.fraunhofer.de} \\
  \And
 Alexander Sauer \\
  Fraunhofer Institute for Manufacturing Engineering and Automation IPA,\\
  Nobelstraße 12, Stuttgart, 70569, Germany\\
  ORCID: 0000-0003-3822-1514 \\
}

\begin{document}

\twocolumn[
    \begin{@twocolumnfalse}
        \maketitle
        \textsuperscript{1} equal contribution \\
        \textsuperscript{*} corresponding author\\
        \begin{abstract}
                        The increasing integration of renewable energy sources has led to greater volatility and unpredictability in electricity generation, posing challenges to grid stability. Ancillary service markets, such as the German control reserve market, allow industrial consumers and producers to offer flexibility in their power consumption or generation, contributing to grid stability while earning additional income. However, many participants use simple bidding strategies that may not maximize their revenues. This paper presents a methodology for forecasting bidding prices in pay-as-bid ancillary service markets, focusing on the German control reserve market. We evaluate various machine learning models, including Support Vector Regression, Decision Trees, and k-Nearest Neighbors, and compare their performance against benchmark models. To address the asymmetry in the revenue function of pay-as-bid markets, we introduce an offset adjustment technique that enhances the practical applicability of the forecasting models. Our analysis demonstrates that the proposed approach improves potential revenues by 27.43\,\% to 37.31\,\% compared to baseline models. When analyzing the relationship between the model forecasting errors and the revenue, a negative correlation is measured for three markets; according to the results, a reduction of 1€\,/MW model price forecasting error (MAE) statistically leads to a yearly revenue increase between 483\,€/MW and 3,631\,€/MW. The proposed methodology enables industrial participants to optimize their bidding strategies, leading to increased earnings and contributing to the efficiency and stability of the electrical grid.
                        \newline
                        \newline
        \end{abstract}
        \vspace{1cm}
    \end{@twocolumnfalse}
]

\section{Introduction}
The global transition towards renewable energy sources, such as wind and solar power, has significantly increased the volatility and unpredictability of electricity generation.
A related study~\cite{ortner2019future} reveals that the number of grid imbalances strongly relates to a country's renewable energy share. These renewable sources are inherently intermittent, leading to fluctuations that can challenge the stability of electrical grids.

In these ancillary service markets, industrial consumers and producers can participate by offering flexibility in their power consumption or generation capacities. They can provide balancing services by adjusting their power usage or supplying balancing energy. By participating in these markets, industrial entities not only contribute to grid stability but also have the opportunity to earn additional income.

Currently, many participants in the ancillary service markets employ simple bidding strategies, which may not optimally reflect market conditions or maximize their potential revenues. Accurate forecasting of bidding prices is essential for developing effective bidding strategies that can enhance earnings while ensuring reliable grid operations.

This paper focuses on developing a robust methodology for forecasting bidding prices in ancillary service markets, specifically targeting the German control reserve market (\textit{Regelenergiemarkt}). By improving price forecasts, industrial consumers can optimize their participation in these markets, thereby increasing their revenues and contributing to the efficiency of the grid.

\subsection{Related Work}

Balancing energy provision is a crucial option for marketing energy flexibility. The guideline 5207 \cite{Blatt1VDI2020} on industrial flexibility marketing (specified in \cite[p.~37-39]{tristan2024methodology}) explicitly lists control reserve markets. The problem of forecasting control reserve market prices falls under multi-step time series forecasting. Several related works align with this theme and are discussed below.

\subsubsection*{Related Forecasts}

There is extensive literature on forecasting aggregated electricity consumption (\cite{poplawski2015forecasting, wang2018ensemble, borges2012evaluating, nazir2023forecasting}) and generation, such as photovoltaic \cite{raza2016recent, rahimi2023comprehensive} or wind power \cite{simankov2023review}, as well as their integrated combination as imbalance forecasts (\cite{beichter2022net, markovics2024short}). 
However, this paper focuses on price forecasting in the control reserve market.

Despite significant literature on price forecasting in day-ahead markets, balancing markets have received less attention \cite{klaeboe2015benchmarking, merten2020automatic}.
To our knowledge, there is no comprehensive survey dedicated to this topic.

\subsubsection*{Control Reserve Market: Analysis, Optimization, and Forecasting}

Li et al. \cite{10161755} investigate the statistical properties of \ac{mFRR} markets in Croatia, France, and Germany, which have different market designs.
Their findings indicate that no single model can describe all \ac{mFRR} markets in Europe. For the German market, no correlation with other studied markets was found, making forecasting more challenging.

Navarro \cite{bassy2024viability} provides a detailed exploratory analysis of the German balancing energy market.

Bezold et al. \cite{Bezold2024-tp} examine the profitability of flexibility in control reserve markets. The study focuses on determining an optimal bid price when held constant over a certain period. 
The proposed method calculates this price and tests it across four German markets to quantify the marketing potential.

Narajewski and Zielewicz \cite{narajewski2022probabilistic} forecast current balancing energy costs based on already published \ac{aFRR} and \ac{mFRR} prices, providing probabilistic predictions for the German balancing market.

Hirth and Ziegenhagen \cite{hirth2015balancing} present an overview of balancing energy market types, sizes, designs, and the European context.

Merten et al. \cite{merten2020automatic} review previous work on the German \ac{aFRR} market. 
In their work, they forecast the market using the models Exponential Smoothing, SARIMA, neural networks, and recurrent networks, but no universally best model was identified.
Incorporating exogenous data sources such as weather, electrical loads or other market data did not improve forecasting performance.
Different from our work, however, no analysis of the \ac{mFRR} markets took place. 
Also, only the forecasting metrics (statistically) were calculated, but no economic consideration or optimization was made.  
Lastly, the values are not longer relevant, as the market structure has changed in the meantime~\cite{regelleistung_historie}. 

Lucas et al. \cite{lucas2020price} forecast prices in the UK pay-as-bid balancing market using Random Forest, Gradient Boosting, and XGBoost, incorporating several exogenous variables. 
They claim that price peaks are, in principle, hard to predict, as prices may result from complete bidding strategies.  
Feature importance analysis showed that Net Imbalance Volume, de-rated margins, and month indicators had the most significant influence. In their setup, Gradient Boosting performed best. 

Vandezande et al. \cite{vandezande2010well} discuss balancing market designs in Europe, focusing on the future expansion of wind energy and advocating for cross-border markets. They address the connection between the limited predictability of external indicators, such as the electricity generation from wind, and the resulting influence on the amount of balancing power required. The latter, in turn, influences the price.

\subsection{Scope and Contributions}

This paper develops a methodology for forecasting bidding prices in pay-as-bid ancillary service markets, focusing on the German control reserve market. Accurate bidding price forecasts are essential for developing effective strategies that enhance earnings for industrial participants while supporting grid stability.

The main contributions of this paper are:

\begin{itemize}
    \item We propose a methodology for day-ahead price forecasting in the German control reserve market, tailored to the characteristics of pay-as-bid markets.
    \item We evaluate various machine learning models for price forecasting, including Support Vector Regression, Decision Trees, and k-Nearest Neighbors, comparing their performance against benchmark models.
    \item We introduce an offset adjustment technique to address the asymmetry in the revenue function of pay-as-bid markets, enhancing the forecasting models' practical applicability.
    \item We provide an analysis of the forecasting results, demonstrating improvements in revenue compared to baseline benchmarks, and discuss the implications for industrial participants.
\end{itemize}

This work is based on three key assumptions.
\begin{itemize}
    \item First, we assume a price-taking setting, i.e., our bidding strategy does not change the pricing or volume of the markets.
    \item Second, the market dynamics are assumed to remain largely stable over time.
    \item Third, since these markets are relatively small and not yet highly liquid, they exhibit partial inefficiencies, creating opportunities to profit from price discrepancies.
\end{itemize}

The remainder of this paper is organized as follows: Section~\ref{sec:methodology} describes the proposed methodology, including model selection and evaluation criteria. Section~\ref{sec:experimental_setup} details the experimental setup and data used. Section~\ref{sec:results} presents the results and analysis. Section~\ref{sec:Outlook} discusses potential improvements and future work. Finally, Section~\ref{sec:conclusion} concludes the paper.

\section{Methodology}
\label{sec:methodology}
\begin{figure*}[htbp]
	\centering
	\includegraphics[width=.95\textwidth]{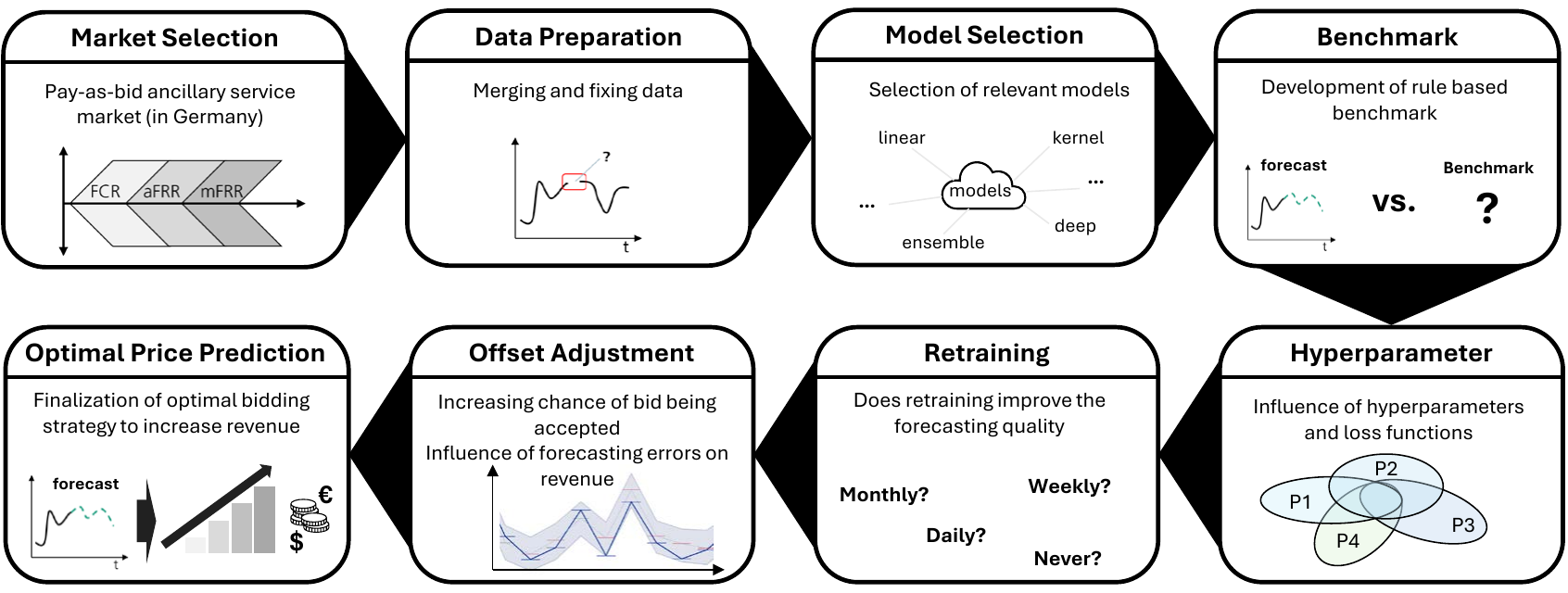}
	\caption{Graphical overview of the methodology.}
	\label{fig:graphical_overview}
\end{figure*}

To identify the optimal bidding strategy for an ancillary services market, we propose the following methodology to develop a robust bidding forecast. 
This methodology, illustrated in Figure~\ref{fig:graphical_overview}, begins with selecting the targeted market and culminates in optimal price predictions. The process is divided into eight distinct modules, each comprehensively detailed in Sections~\ref{sec:market_selection} through \ref{sec:output_price_prediction}. 

\subsection{Market Selection}
\label{sec:market_selection}
The selection of appropriate markets employing pay-as-bid mechanisms with day-ahead pricing is crucial for applying of the proposed method. Ancillary service markets are typically divided into several submarkets, each catering to different response times and specific characteristics. Generally, the markets for capacity provision, where power capacity is reserved, are separate from those where energy is dispatched and provided. This method focuses exclusively on the control reserve market (\textit{Regelleistungsmarkt}), assuming that the second part of the bid in the balancing energy market (\textit{Regelarbeitsmarkt}) is designed to cover the provider's operational costs. Depending on the submarket, various mechanisms and products are utilized. 
For the effective implementation of this method, it is crucial that the market operates on a pay-as-bid basis.

\subsection{Data Preparation}
Data collection is a fundamental prerequisite for the implementation of the proposed methodology.
Ancillary service market data are typically obtained from the websites of transmission system operators or through \ac{API} requests. 
From the acquired datasets, the relevant variables and parameters must be meticulously extracted for analysis.
It is essential to thoroughly examine the data to identify and eliminate any irregularities or inconsistencies.

\subsection{Benchmark}
\label{sec:benchmark}
To evaluate the proposed method and selected models, a benchmark is necessary for comparison. Simple rule-based benchmarks, such as a naive forecast using the previous day's bidding values, serve as a reference point. 
These basic models provide a baseline, allowing for the assessment of how more advanced methods improve upon straightforward forecasting strategies.

\subsection{Model Selection}
\label{sec:model_selection}
In selecting appropriate forecasting methods, various classes of algorithms can be considered, each with distinct characteristics suited to different requirements.
Linear models, such as linear regression, are computationally efficient~\cite{Diakonikolas2018Efficient, Tay2021Elastic} and straightforward to implement~\cite{James2021Moving}; however, they are limited in their ability to capture complex nonlinear relationships within the data~\cite{Andersen2009Nonparametric}.
Kernel-based methods like \acfp{SVM} are effective for modeling nonlinear patterns~\cite{Zeng2019Multiple} and perform well with smaller datasets~\cite{Nguyen2022Robust}, but they can become computationally intensive with larger data volumes~\cite{Nguyen2022Robust}. 
Ensemble models, including techniques like Random Forests and Gradient Boosting Machines, offer a balanced compromise between computational speed, generalization ability, and practical applicability\cite{Bentejac2019A}, making them suitable for a wide range of forecasting tasks. 
Deep learning models, particularly \acp{RNN} or \ac{LSTM} networks, excel at capturing intricate nonlinear dependencies and have strong generalization capabilities~\cite{Neyshabur2017Exploring}. 
Nevertheless, they typically require large amounts of training data~\cite{Whang2020Data} and substantial computational resources~\cite{Shen2024On}, which may limit their feasibility in scenarios with limited data or computational constraints. Other model classes and adaptations tailored to specific tasks can also be considered to further enhance forecasting performance.

\subsection{Hyperparameter}
\label{sec:hyperparameter}
In this module, the pipeline is extended by the functionality of hyperparameter optimization. 
The hyperparameter optimization method and the loss function used for training are selected. 
The influence of both choices can then be systematically analyzed.

\subsection{Retraining}
\label{sec:retraining}
\begin{figure}[h]
	\centering
	\includegraphics[width=.9\columnwidth]{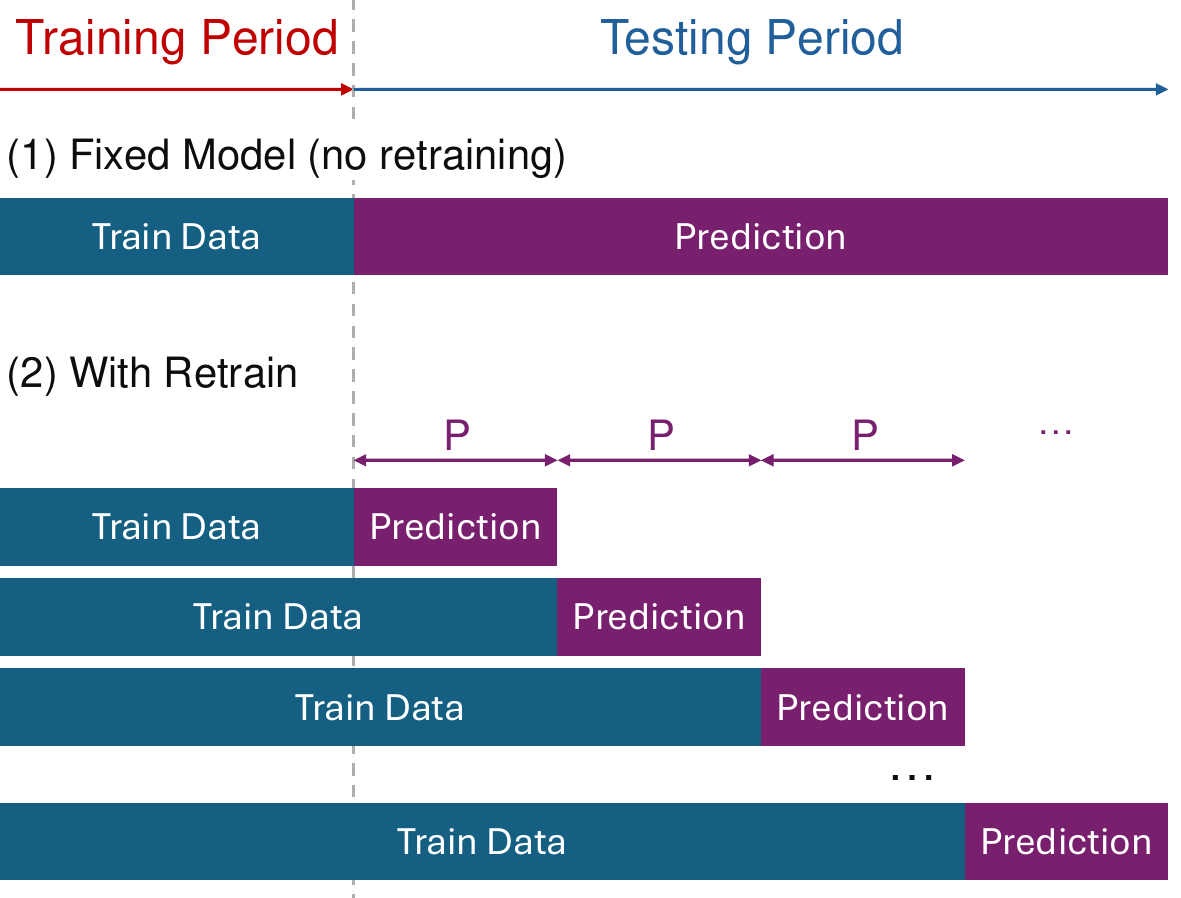}
	\caption{Schematic representation of retraining.}
	\label{fig:retraining}
\end{figure}

As the market contains comparatively few data points (typically a few per day) and volatility constantly changes, the model must be retrained occasionally.
To measure this effect, the performance of a model trained once on the training data and then kept constant is compared with models regularly retrained with the new data from the evaluation period.
This concept is illustrated in Figure~\ref{fig:retraining}.
In the first case (1), the model is kept fix over the whole evaluation period, while in the other case (2), the model is retrained every time interval $P$, using all data that is available till then.
By doing so, short-term market changes can be handled with the model.

\subsection{Offset Adjustment}
\label{sec:offset_adjustment}
By training the models, an attempt is made to reduce errors, i.e., the spread of the residual between predicted and actual prices.
Model optimization using loss is typically symmetrical, this results in the errors being relatively evenly distributed. 

\begin{figure}[h]
	\centering
	\includegraphics[width=\columnwidth]{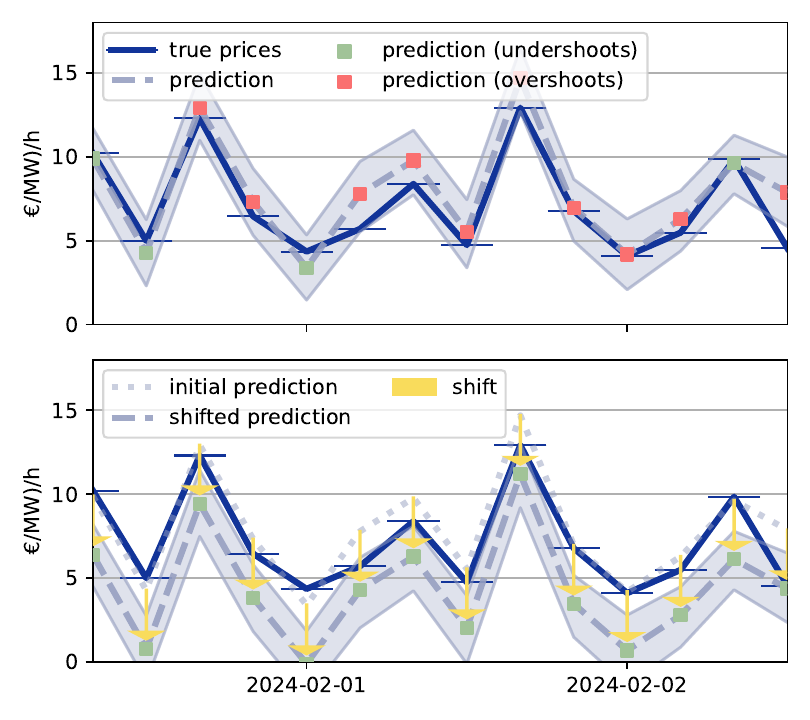}
	\caption{Comparison of forecast without (top) and with offset adjustment (bottom).}
	\label{fig:offset_adujstment}
\end{figure}

The revenue function in the pay-as-bid scenario, on the other hand, is asymmetrical: if the prediction exactly matches the actual price, the revenue is maximum. 
For predictions below the actual price, the difference between price and prediction corresponds exactly to the revenue deviation.
For predictions above the actual price, however, the revenue is 0.  
Accordingly, overshoots are worse than undershoots.
Figure~\ref{fig:offset_adujstment}~(top) illustrates this: a prediction and its symmetric prediction interval are plotted against the true price curve. 
According to this, some forecasts are higher, and some lower than the actual prices, but only undershoots will benefit the revenue.

Offset adjustment corrects the forecast $\hat{y}$ downwards by a fixed offset $\delta$. 
As a result, fewer higher prices are issued overall, but the number of times awarded increases. 
The value 
\begin{align}
        o^* =  \underset{\delta}{\argmax} \ \texttt{revenue}(y, \hat{y} + \delta)
\end{align}

is obtained by shifting the forecasts on a reference time window until a maximum is reached.
Figure~\ref{fig:offset_adujstment}~(bottom) illustrates the shifted prediction by an offset. 
Compared to the original forecast, more forecasts lie below the actual price curve $y$. 
The details for determining a suitable $\delta$, as well as the influence of the forecasting error on the resulting revenue in general, are explained based on the concrete markets later in Section~\ref{sec:result_offset_adjustment}.   

\subsection{Output: Price Prediction}
\label{sec:output_price_prediction}
The output of the pipeline is the actual, offset-optimized prediction.
It is directly used as day-ahead price.

\section{Experimental Setup}
\label{sec:experimental_setup}
This section details the abstract framework from the previous Section~\ref{sec:methodology}. 
All listed steps are sequentially described, based on the German market case study.

\subsection{Market}
\label{sec:market}

\begin{figure}[h]
	\centering
	\includegraphics[width=1.0\columnwidth]{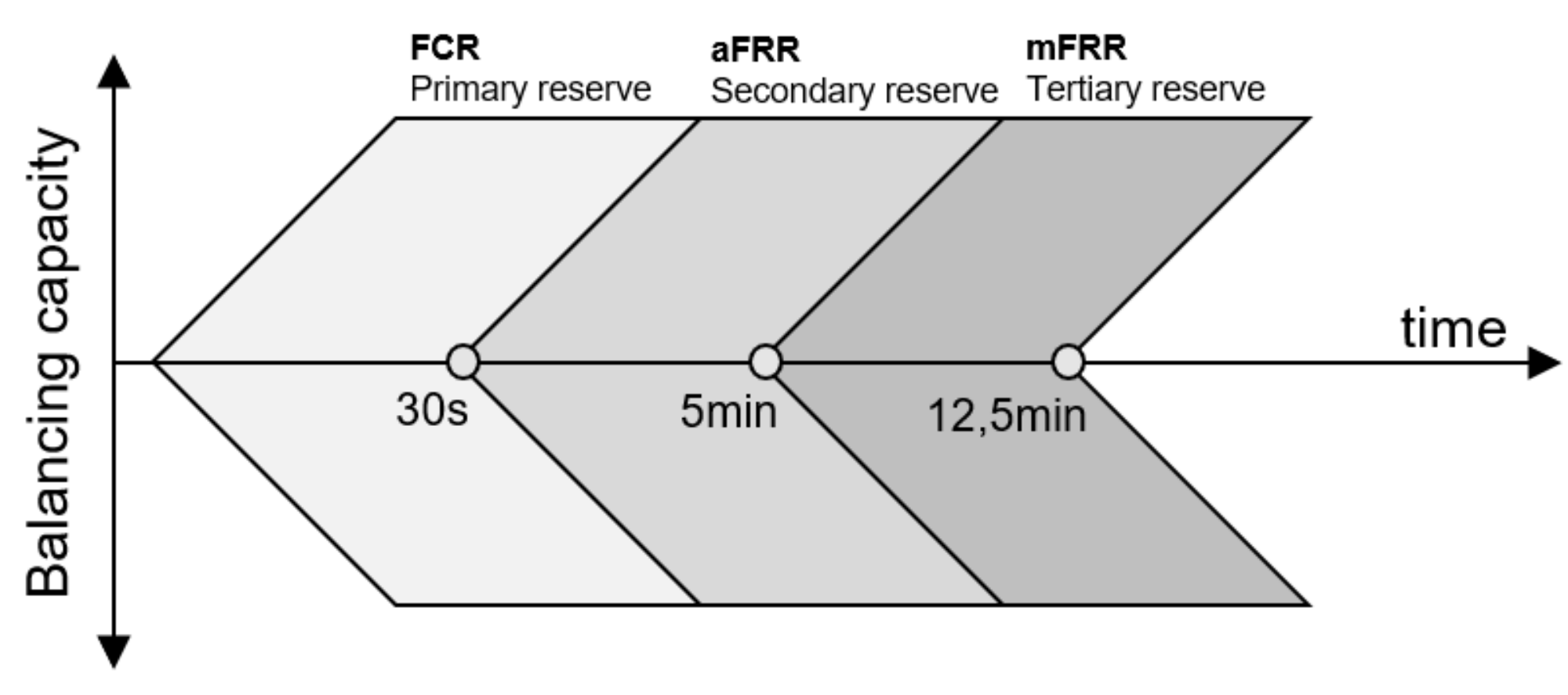}
	\caption{German control reserve market and its submarkets. (based on \cite{nextkraftwerke_regelenergie})}
	\label{fig:RegelleistungsmarktSchema}
\end{figure}
Like most markets, the German control reserve market is divided into different submarkets. 
The primary differentiation is based on the time the control reserve must be available (see Figure~\ref{fig:RegelleistungsmarktSchema}). The primary reserve, also known as Frequency Containment Reserve (\ac{FCR}), must be available within 30 seconds~\cite{regelleistung_FCR}; the secondary reserve (Automatic Frequency Restoration Reserve, \ac{aFRR}) within 5 minutes~\cite{regelleistung_aFRR}; and the tertiary reserve (Manual Frequency Restoration Reserve, \ac{mFRR}) within 12.5 minutes~\cite{regelleistung_mFRR}. 
Since the response time for \ac{FCR} is very short, it is predominantly provided by batteries. Additionally, the \ac{FCR} market is not split into capacity and energy markets, because its goal is continuously balancing the grid. In contrast, the \ac{aFRR} and \ac{mFRR} markets are divided into capacity and energy markets. The energy market operates on a pay-as-cleared basis, whereas the capacity market operates on a pay-as-bid basis. As the proposed method (see Section~\ref{sec:methodology}) is only valid for pay-as-bid markets, it is applied solely to the capacity market. Moreover, these markets are further split into positive and negative balancing power, resulting in four markets each for capacity and energy. This work will examine the four capacity markets, as the method is applicable only to the capacity market. Additional market-specific characteristics can be found in Table~\ref{tab:requirements}.
The auction for \ac{aFRR} closes at 9 a.m., and for \ac{mFRR} at 10 a.m., on the day before procurement \cite{regelleistungBeschaffung}. The entire following day is traded, divided into six blocks of four hours each. The historical data can be downloaded from the official website regelleistung.net~\cite{regelleistungStartseite}.

\begin{table*}[htbp]
\centering
\small
\caption{Overview of FCR, aFRR, and mFRR requirements \cite{ertz2022}.}
\begin{tabularx}{\textwidth}{|X|X|X|X|}
\hline
\textbf{Requirement} & \textbf{FCR} & \textbf{aFRR} & \textbf{mFRR} \\ 
\hline
Period per incident to be covered & $0 \leq t \leq 15$ min & $30 \, \text{s} \leq t \leq 15$ min & $15 \, \text{min} \leq t \leq 60$ min \\ 
\hline
Permissible response time & $\leq 30 \, \text{s}$ & $\leq 5 \, \text{min}$ & $\leq 12.5 \, \text{min}$ \\ 
\hline
Product length & 4h & 4h & 4h \\ 
\hline
Minimum Capacity & $\pm 1 \, \text{MW}$ (symmetric) & $1 \, \text{MW}$ (positive or negative) & $1 \, \text{MW}$ (positive or negative) \\ 
\hline
Minimal working capacity & $2 \cdot t_{\text{max}} \cdot P$ & $t_{\text{max}} \cdot P$ & $t_{\text{max}} \cdot P$ \\ 
\hline
Control & Fully automatic & Fully automatic & Manual \\ 
\hline
Payment modality & pay-as-cleared  & Capacity price (pay-as-bid),  Power price (pay-as-cleared)  &  Capacity price (pay-as-bid),  Power price (pay-as-cleared) \\ 
\hline
\end{tabularx}
\label{tab:requirements}
\end{table*}

\subsection{Data and Analysis}
\label{sec:data_and_setup}
\begin{figure}[h]
	\centering
	\includegraphics[width=1.0\columnwidth]{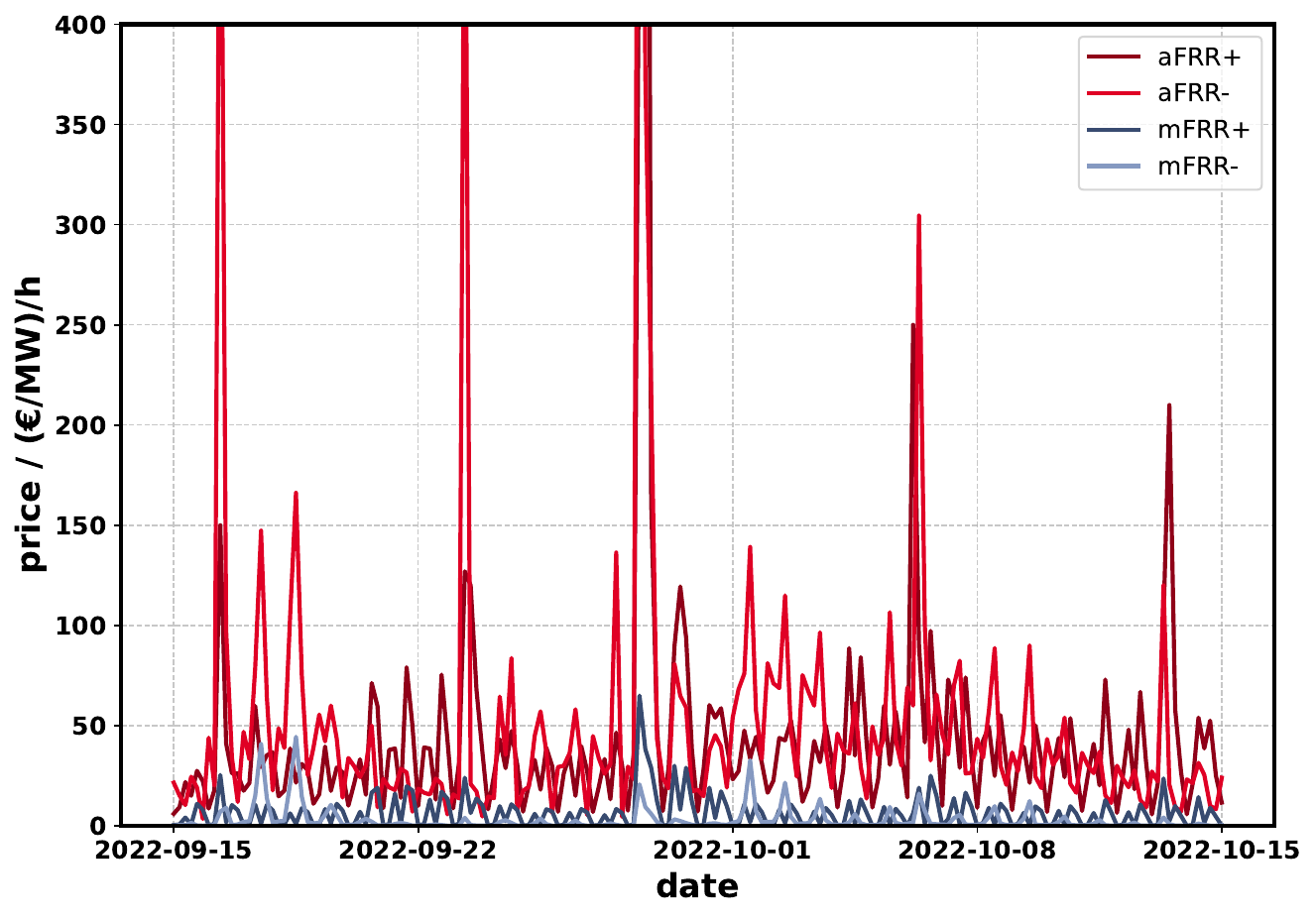}
	\caption{A sample of data for the different markets is shown.}
	\label{fig:sampleData}
\end{figure}

The data for this study was downloaded from~\cite{regelleistung_Download}. 
As a visualization of the markets and the bidding values, a sample of the data is provided in Figure~\ref{fig:sampleData}. 
Since the market mechanism was changed in 2020~\cite{regelleistung_historie}, only data from January 1, 2021, to June 30th, 2024 was considered. The data was split into training and validation sets, with the validation period covering data from 2024. 
This results in approximately three years of training data and half a year of validation data, representing a split of roughly \SI{85.7}{\%} for training and \SI{14.3}{\%} for validation.

The grid operators directly provide the data. They are the official market data platform. 
Accordingly, no data cleaning or preparation was necessary, as there were no missing values or significant outliers.

As part of the preliminary analysis, the dataset's basic statistics were calculated and presented in Table~\ref{tab:basicMetrics}. The analysis reveals differences between the markets. For instance, the mean value of the \ac{aFRR} market is, on average, three times higher than that of the \ac{mFRR} market, which can be attributed to the higher response time requirements of \ac{aFRR}.

\begin{figure}[h]
	\centering
	\includegraphics[width=1.0\columnwidth]{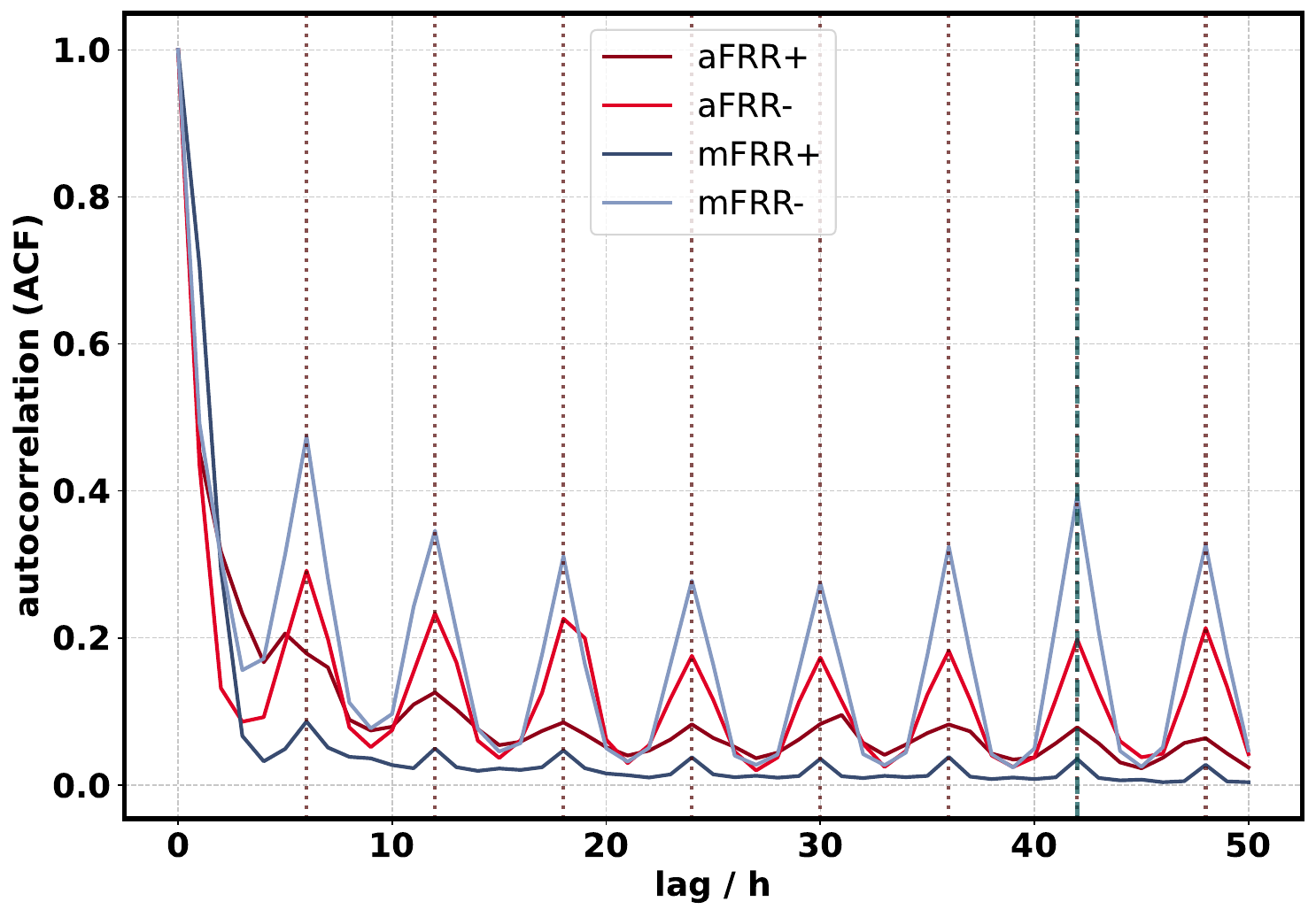}
	\caption{Autocorrelation of the four investigated markets. Lag 6 (\SI{24}{h}) is marked with dotted lines. Lag 42 (\SI{1}{w}) is marked separately with another type of dashed line.}
	\label{fig:autocorrelation}
\end{figure}

Additionally, an autocorrelation analysis was conducted (see Figure~\ref{fig:autocorrelation}). 
The analysis shows strong autocorrelations on a daily basis (with one lag representing 4 hours) and a slight increase in autocorrelation on a weekly basis. 
Based on these observations, a benchmark bidding method is derived that is further discussed in Subsection~\ref{sec:baselines}.

\begin{table}[h]
\centering
\caption{Summary statistics for the German \ac{aFRR} and \ac{mFRR} markets for the training dataset from 2021 to 2023.}
\begin{tabular}{lrrrr}
\hline
\SI{}{\euroMWh} & \textbf{\ac{aFRR}+} & \textbf{\ac{aFRR}--} & \textbf{\ac{mFRR}+} & \textbf{\ac{mFRR}--} \\ 
\hline
\textbf{Mean}  & 21.36 & 24.38 & 7.99 & 7.35 \\ 
\textbf{Std}  & 72.14 & 70.13 & 43.81 & 21.67 \\ 
\textbf{Min} & 0.50 & 0.30 & 0.00 & 0.00 \\ 
\textbf{Max} & 1871.04 & 1979.0 & 2437.5 & 798.77 \\ 
\textbf{Q25}  & 6.42 & 3.77 & 1.00 & 0.35 \\ 
\textbf{Median} & 10.85 & 9.21 & 2.99 & 1.33 \\ 
\textbf{Q75} & 19.67 & 23.14 & 8.52 & 5.34 \\ 
\hline
\end{tabular}
\label{tab:basicMetrics}
\end{table}

\begin{table}[h!]
\centering
\caption{Summary statistics for the German \ac{aFRR} and \ac{mFRR} markets for the test dataset of the first 6 months of 2024.}
\begin{tabular}{lrrrr}
\hline
\SI{}{\euroMWh} & \textbf{\ac{aFRR}+} & \textbf{\ac{aFRR}--} & \textbf{\ac{mFRR}+} & \textbf{\ac{mFRR}--} \\ 
\hline
\textbf{Mean} & 15.89 & 13.95 & 3.82 & 10.87 \\ 
\textbf{Std} & 41.25 & 21.67 & 6.20 & 17.72 \\ 
\textbf{Min} & 2.11 & 1.09 & 0.08 & 0.10 \\ 
\textbf{Max} & 992.34 & 285.97 & 69.00 & 170.00 \\ 
\textbf{Q25} & 5.00 & 3.76 & 0.53 & 2.00 \\ 
\textbf{Median} & 8.82 & 6.51 & 1.34 & 3.62 \\ 
\textbf{Q75} & 16.97 & 15.16 & 4.43 & 11.05 \\ 
\hline
\end{tabular}
\label{tab:basicmetricsTest}
\end{table}

\subsection{Forecasting Setup}
\label{sec:forecasting setup}
The forecasting setup was multi-shot univariate with a forecasting horizon of 24 hours for all markets.
As one forecast period lasts four hours, six points have to be forecasted at once daily.
Motivated by a previously conducted autocorrelation study, all models use a historical input horizon of 42 points, i.e., one week.

\subsection{Models}
\label{sec:models}
Forecasting can be accomplished using a wide range of models; for instance, scikit-learn~\cite{scikit-learn} provides numerous regression algorithms suitable for this purpose. 
However, due to limitations in computational resources and the redundancy of models with similar characteristics, it is practical to focus on a selected subset of models. 
Therefore, we have chosen to include at least one model from each of the most common classes of regression algorithms to cover a broad spectrum of modeling approaches. 
Given the computational constraints, a pre-selection was conducted, and neural network-based models were excluded. 
Moreover, several sources \cite{toner2024analysislineartimeseries, green2015simple} suggest that, particularly in time series forecasting, simpler models often outperform more complex ones.
This aspect warrants further investigation in future work. The models and the model classes are listed in Table~\ref{table:models}.
\begin{table}[h!]
\centering
\caption{Classification of models by type.}
\begin{tabularx}{\columnwidth}{|l|X|}
\hline
\textbf{Model} & \textbf{Type} \\ \hline
\acf{SVR} & Support Vector Machine \\ \hline
\acf{kNNR} & K-Nearest Neighbors \\ \hline
\acf{DTR} & Decision Tree \\ \hline
RandomForestRegressor & Ensemble (Bagging) \\ \hline
GradientBoostingRegressor & Ensemble (Boosting) \\ \hline
\acf{XGB} & Ensemble (Boosting) \\ \hline
LGBMRegressor & Ensemble (Boosting) \\ \hline
Ridge & Linear Model (Regularized) \\ \hline
Lasso & Linear Model (Regularized) \\ \hline
ElasticNet & Linear Model (Regularized) \\ \hline
\end{tabularx}
\label{table:models}
\end{table}

\subsection{Baselines}
\label{sec:baselines}
As described in Subsection~\ref{sec:benchmark}, establishing a robust and meaningful benchmark is critical for comparing the performance of various models. To achieve this, we implemented two distinct classes of benchmark models: the \textit{Fixed Bid Benchmark} and the \textit{Lagged Forecast Benchmark}, which are explained in the remaining of this subsection.

\subsubsection{Fixed Bid Benchmark}
\label{sec:fixed_bid_benchmark}
The \textit{Fixed Bid} benchmark relies on a static bid value determined over a predefined period. For instance, the bid value is chosen as the one that would have maximized revenue during a specific historical period, such as the previous day, week, or month. This approach was originally proposed by Bezold et al.~\cite{Bezold2024-tp}.

\subsubsection{Lagged Forecast Benchmark}
\label{sec:lagged_forecast_benchmark}
The \textit{Lagged Forecast} benchmark is informed by the autocorrelation analysis presented in Figure~\ref{fig:autocorrelation}. The analysis reveals significant correlations at both daily and weekly intervals, with daily correlations being more pronounced. Based on these insights, we developed two benchmark models where the predicted values are directly taken from either the previous day or the previous week. These models are collectively referred to as the \textit{Lagged Forecast} benchmark.

\subsection{Loss Functions}
\label{sec:loss_functions}
Three loss functions were selected as a measure of the performance of the models.
The performance of each model is examined based on the predicted values $\hat{y}_i$ and the actual observed prices $y_i$ for each hour $i \in T$.

Due to its high popularity\cite{nti2020electricity}, \ac{MAPE} was chosen as a relative, and \ac{MAE} as an absolute metric:
\begin{align}
    \texttt{MAE} (y, \hat{y})  & = \frac{1}{|T|}\sum_{i \in T} | \hat{y}_i - y_i |\\
    \texttt{MAPE} (y, \hat{y}) & = \frac{1}{|T|}\sum_{i \in T} \frac{| \hat{y}_i - y_i |}{y_i}
\end{align}

In addition to the deviation of the curves, the economic performance of the forecast for a period can also be determined directly by calculating the surcharged prices and the associated incremented revenues directly. This function is given as
\begin{align}
\texttt{revenue}(y, \hat{y}) &= \sum_{i \in T} \hat{y}_i \cdot \mathds{1}_{\hat{y}_i \leq y_i}
\end{align}
where $\mathds{1}$ is the indicator function, evaluating to $1$ for predictions being smaller or equal to the actual price (there is a surcharge), or $0$ otherwise.

\subsection{Hyperparameter}
\label{sec:hyperparamter}

As a hyperparameter optimization algorithm, grid search with $k$-fold cross-validation was chosen for our experiments to ensure model generalizability even on few data. 
In all experiments, $k=5$ was used, since it is a typical value in literature\cite{nti2021performance}. 
This is a reasonable compromise between generalizability of the hyperparameter evaluation and computational effort. 
The parameter search spaces used for tuning the hyperparameters are listed in the Appendix~\ref{sec:used_grid_search_parameters}.

All three loss functions, as defined in Subsection~\ref{sec:loss_functions}, were examined for hyperparameter optimization. 
An in-depth analysis of the influence of hyperparameter optimization on prediction performance can be found in the results section (Subsection~\ref{sec:influence_hyperparametertuning_and_loss_function}).

\section{Results}
\label{sec:results}
This section is divided into three parts: the first section (Subsection~\ref{sec:influence_hyperparametertuning_and_loss_function}) compares the machine learning models with the benchmarks and analyzes the influence of different hyperparameter tuning settings.
Subsequently (Subsection~\ref{sec:influence_retraining}), the choice of the retraining period on the model performance is analyzed. 
Finally, both the added value of offset adjustment and the influence of model forecast errors on the revenue is evaluated (Subsection~\ref{sec:result_offset_adjustment}). 

\subsection{Model and Benchmark Performances}
\label{sec:influence_hyperparametertuning_and_loss_function}
To improve forecasting quality, a hyperparameter study was conducted. 
The hyperparameters used in the study are listed in Table~\ref{tab:param_grid}. Additionally, different loss functions, as described in Subsection~\ref{sec:loss_functions}, were evaluated. The results for \ac{aFRR}+ are presented in Figure~\ref{fig:models}, while results for other cases are provided in the Appendix~\ref{sec:appendix_hyperparam_images}. Note that the impact of retraining is discussed separately in Subection~\ref{sec:influence_retraining}.

The tentative key findings from the experiment are as follows:
\begin{itemize}
    \item The model choice has the most significant influence on the forecasting quality.
    \item Depending on the market, only three models consistently outperform the benchmark. \ac{SVR} performs best, followed by \ac{DTR} and \ac{kNNR}.
    \item The impact of hyperparameter tuning is marginal for most models and highly dependent on the specific model. In some markets and models, hyperparameter tuning even reduces forecasting quality compared to the default configuration. Overall, the performance improvements from hyperparameter tuning are limited.
    \item The influence of the loss function is generally low. The most stable results are obtained using the negative \texttt{revenue} loss function.
\end{itemize}

In conclusion, both hyperparameter tuning and loss function selection contribute only minimally to improving forecasting quality. 
Consequently, no universal strategy for forecasting improvement can be derived from these experiments. The primary takeaway is that \ac{SVR} is the most suitable model for this task for all markets. 
However, to ensure computational feasibility, only a limited number of hyperparameter configurations were tested in this study (see Section~\ref{sec:appendix_hyperparam_images} for details). 
Future work should explore a broader range of hyperparameter settings for further insights.
  
\begin{figure*}[htbp]
	\centering
	\includegraphics[width=\textwidth]{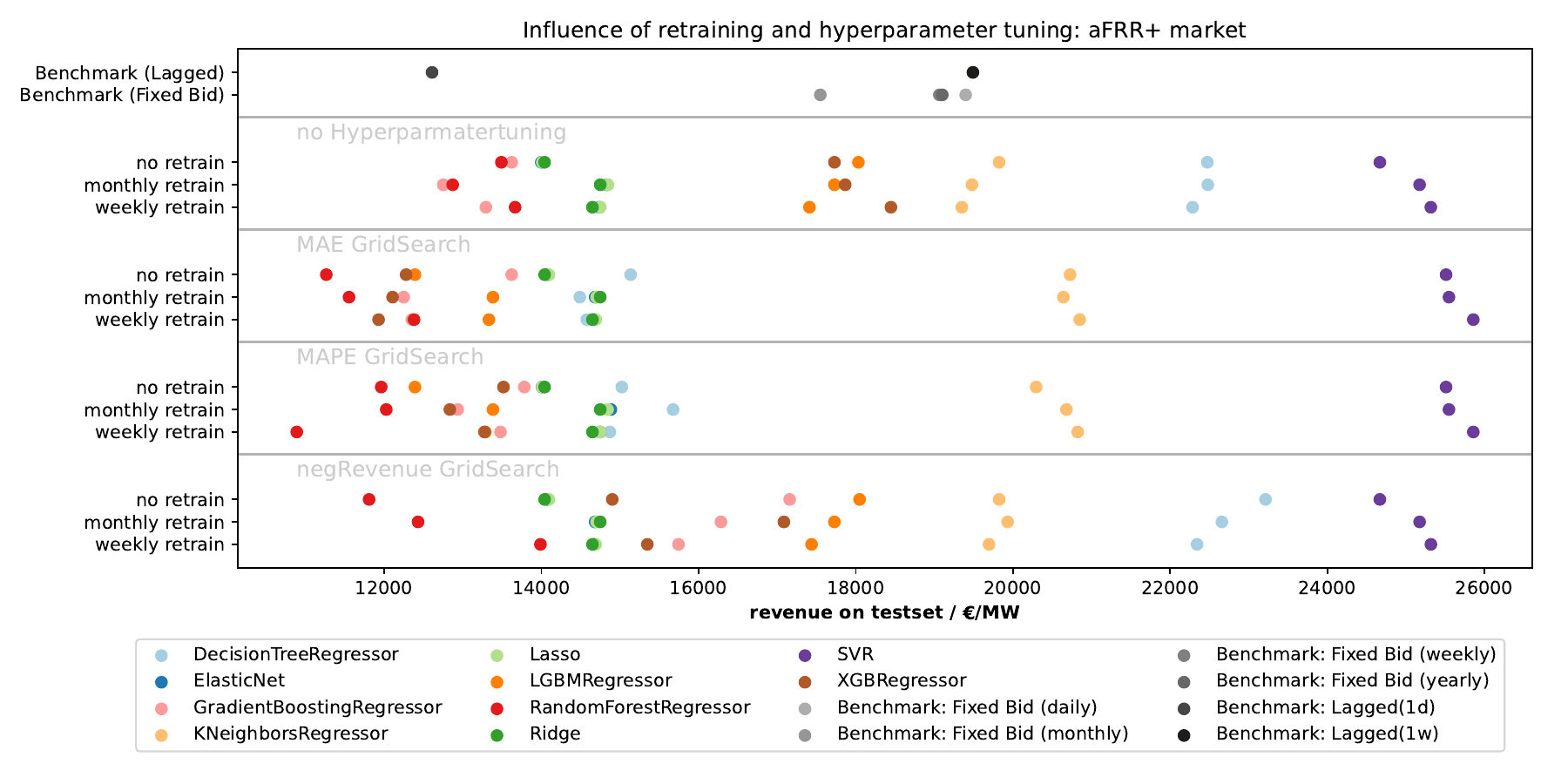}
	\caption{Performance of the models with different hyperparameter tuning loss functions compared to the benchmarks (results for \ac{aFRR}+\ market).}
	\label{fig:models}
\end{figure*}

\subsection{Influence Retraining}
\label{sec:influence_retraining}

To evaluate the impact of retraining on forecasting quality, an analysis was conducted. Since auctions occur daily, the most frequent possible retraining cycle is daily. Due to the inertia of market dynamics, we assume no significant changes occur daily. We also assume an inherent market regularity that permits accurate forecasting. Furthermore, daily retraining requires approximately 30 times more resources than monthly retraining. Consequently, we investigated only the effects of weekly and monthly retraining on forecasting quality. The retraining process adhered to the methodology outlined in Subsection~\ref{sec:retraining}.

As illustrated in Figure~\ref{fig:models} for the \ac{aFRR}+ market (results for other markets are presented in Appendix~\ref{sec:appendix_hyperparam_images}), retraining yields marginal changes in revenue. While some models show a slight improvement in performance, others experience a decrease. This pattern is consistent across all analyzed markets. This behavior can be attributed to the limited increase in the training dataset size, which can only grow by the length of the test dataset, corresponding to one-seventh of the total dataset.

It is noteworthy that retraining substantially increases computational time and resource consumption. Given the insignificant performance improvements, focusing on testing a wider range of models may be more beneficial than emphasizing frequent retraining.

\subsection{Offset Adjustment}
\label{sec:result_offset_adjustment}
\begin{figure*}[htbp]
	\centering
	\includegraphics[width=.95\textwidth]{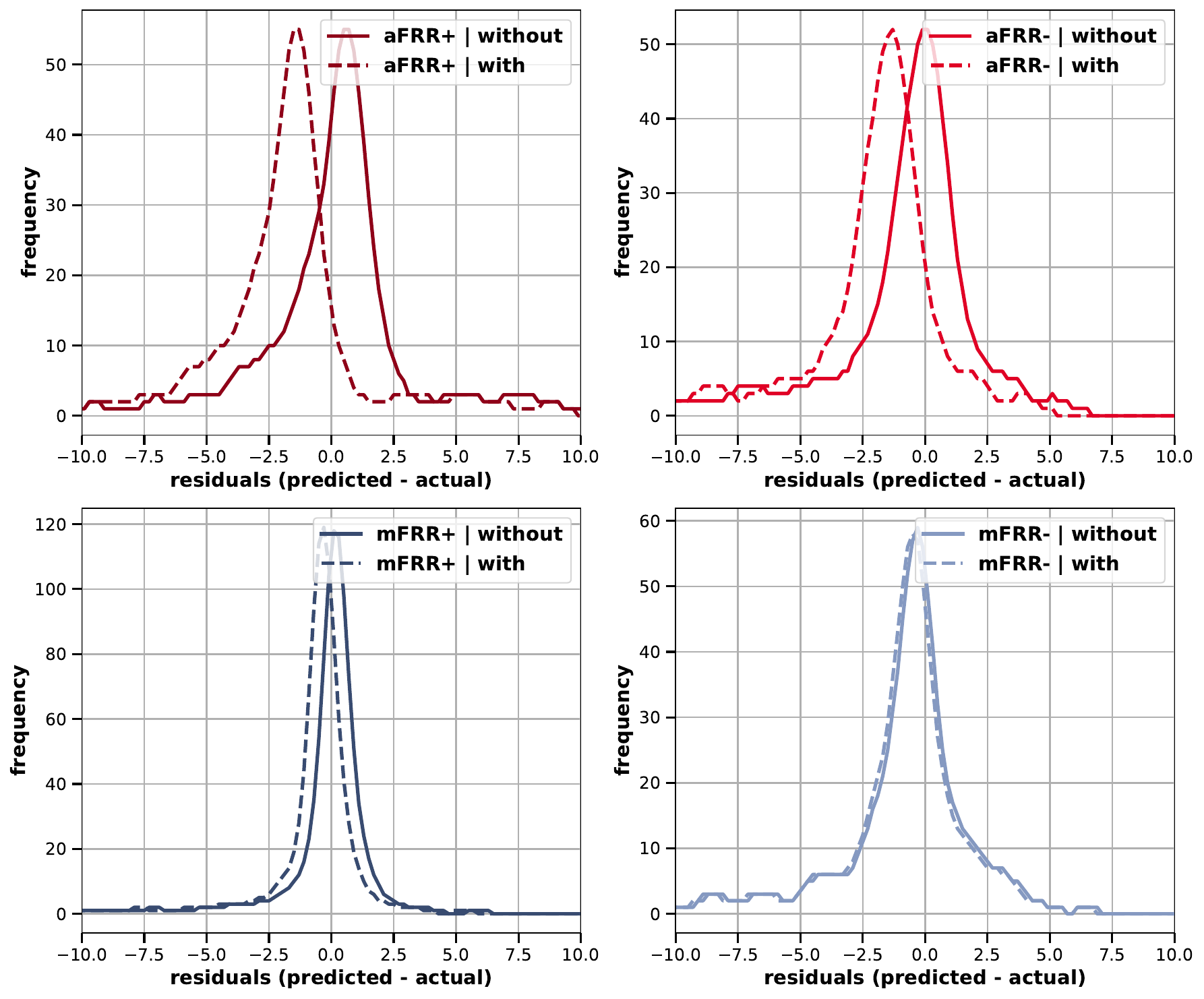}
	\caption{Residuals between predicted and actual values before (without) and after (with) applying offset adjustment using the Support Vector Regressor (SVR) model.}
	\label{fig:SVR_Histogram_Offset_Combined}
\end{figure*}

After successful training, the forecast error of the resulting models ideally scatters symmetrically around 0, i.e., there are roughly as many undershoots as overshoots.
An analysis of the forecast residuals using the example of the \ac{SVR} is shown in Figure~\ref{fig:SVR_Histogram_Offset_Combined}.

However, due to the asymmetry of the revenue function (overshoots are penalized with 0€, undershoots only with the deviation from the real value), it is worth systematically reducing the forecast price.

\subsubsection*{Offset Choice and Improvements}

\begin{figure}[h]
	\centering
	\includegraphics[width=\columnwidth]{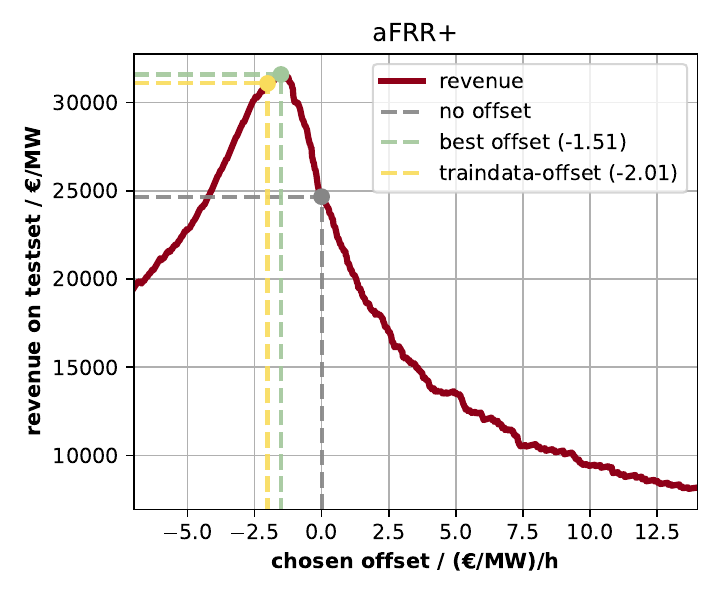}
	\caption{Visualization of the total test data revenue by applying a forecast offset $\delta$ in the \ac{aFRR}+ market using the \ac{SVR} model.}
	\label{fig:offset_adjust__offset_over_revenue}
\end{figure}

Figure~\ref{fig:offset_adjust__offset_over_revenue} illustrates this effect using an \ac{SVR} forecast as an example.
To analyze the effect of choosing an offset $\delta$, different values are used to systematically shift the forecast upward ($\delta>0$) and downward ($\delta < 0$). For each configuration of $\delta$, the resulting revenue is calculated. 
An optimal offset can be calculated ex-post (green line). 
Since this offset is not known in reality, in all our experiments, the offset was determined on the historical training data and then used for the evaluation of the test data (yellow line). 
Compared to the reference forecast (gray, i.e., $\delta=0$), the offset determined in this way is close to the optimum and results in significantly higher revenue than not applying an offset.
The curves of the other models show similar behavior.

\begin{figure*}[htbp]
	\centering
	\includegraphics[width=\textwidth]{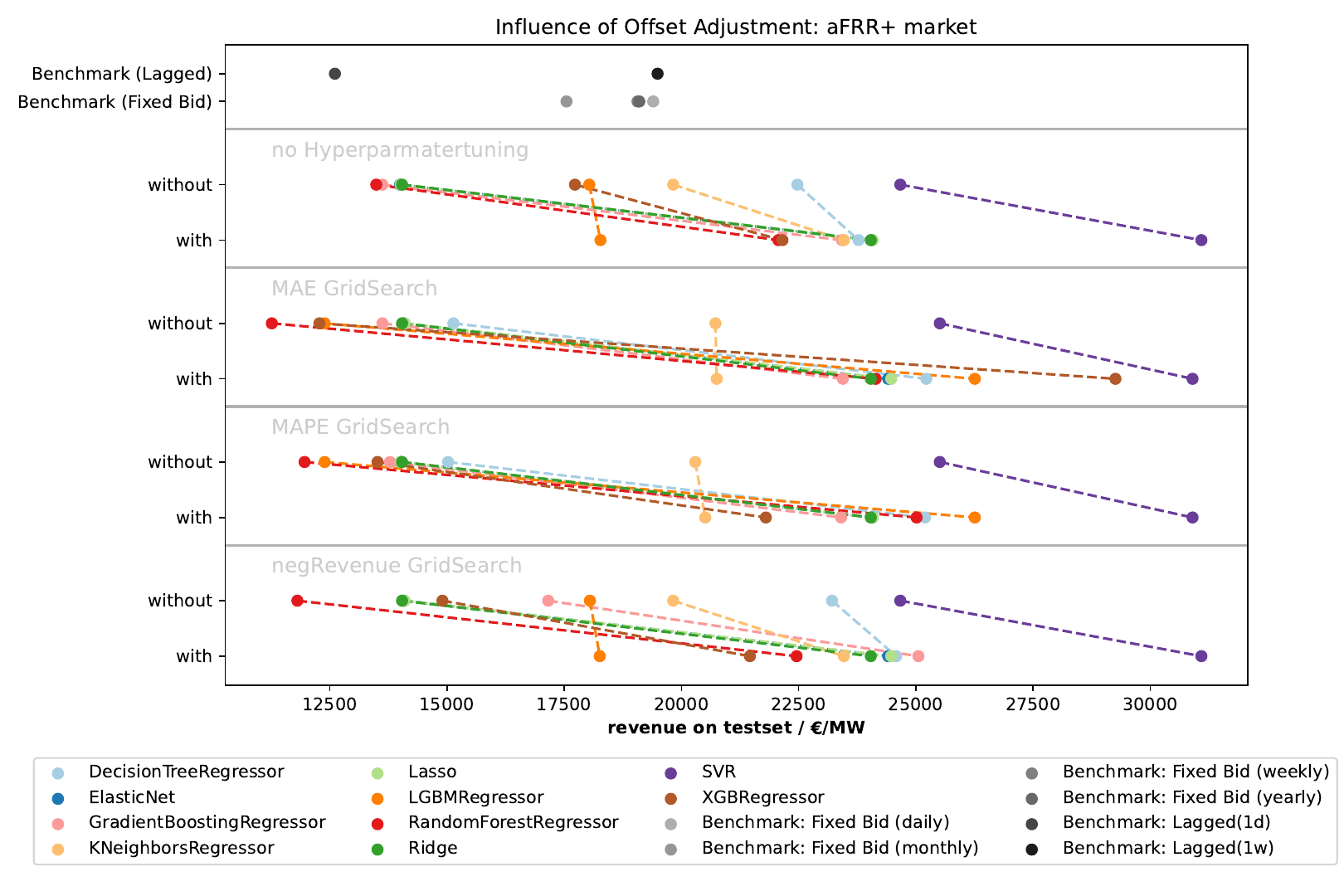}
	\caption{Performance comparison of the models with and without applying Offset Adjustment (results for \ac{aFRR}+\ market).}
	\label{fig:with_and_without_offset_adjustment}
\end{figure*}

The potential that this shift brings for each model class is shown in Figure~\ref{fig:with_and_without_offset_adjustment}. 
It shows the results of the models according to different training modes with and without applied offset adjustment compared to the baseline models in the \ac{aFRR}+~market.
According to this analysis in the \ac{aFRR}+~market, the median model performance increased by 71.2\% (\ac{mFRR}+\ 15.9\% improvement, \ac{aFRR}--\ 25.8\% improvement).
For the \ac{mFRR}--\ market, there was no improvement.

\subsubsection*{Final Results}
Table~\ref{tab:best_scores} shows the results of the final models after offset adjustment and all other optimization steps.
The \ac{SVR} and \ac{XGB} model perform best. Although the error sizes are in different dimensions ($\SI{2.05}{\euroMW}-\SI{7.60}{\euroMW}$~MAE, $\SI{38.27}{\euroMWhsqr}-\SI{1628.52}{\euroMWhsqr}$~MSE), the respective revenues are more dependent on the specific market than on the error. Nevertheless, it can be generalized that the machine learning-based forecast achieves between 27.43\% and 37.31\% more revenue than the market-specific best benchmark in all markets. 
In the test period from January 1, 2024, to June 30, 2024, the revenue from bidding ranged between \SI{5338}{\euroMW} and \SI{31091}{\euroMW}.
The most revenue was generated on the \ac{aFRR} markets, the least for the \ac{mFRR}+.

\begin{table*}[htbp]
\centering
\caption{Summary of the best scores per market evaluated on the testset.}
\begin{tabular}{lrrrr}
\hline
 & \textbf{\ac{aFRR}+} & \textbf{\ac{aFRR}--} & \textbf{\ac{mFRR}+} & \textbf{\ac{mFRR}--} \\ 
\hline
\textbf{Best Model} & SVR & XGBRegressor & XGBRegressor & SVR \\ 
\hline
\textbf{MAE} (\SI{}{\euroMWh}) & 7.60 & 5.88 & 2.05 & 5.05 \\ 
\textbf{MSE} (\SI{}{\euroMWhsqr}) & \num{1628.52} & 291.95 & 38.27 & 136.17 \\ 
\textbf{MAPE} (\SI{}{\%}) & 34.98 & 36.78 & 86.25 & 42.29 \\ 
\hline
\textbf{total revenue, testset} (\SI{}{\euroMW}) & \num{31090.99} & \num{29448.57} & \num{5338.32} & \num{19631.92} \\ 
\textbf{difference wrt. best baseline} (\SI{}{\euroMW}) & \num{+11599.55} & \num{+9976.76} & \num{+2278.41} & \num{+5384.81} \\ 
\textbf{difference wrt. best baseline} (\SI{}{\%}) & +37.31 & +33.88 & +29.91 & 27.43\\ 
\hline
\textbf{total revenue, yearly average} (\SI{}{\euroMW}) & \num{62523.64} & \num{59220.75} & \num{10735.30} & \num{39479.58} \\ 
\textbf{difference wrt. best baseline, yearly average} (\SI{}{\euroMW}) & \num{+23326.57} & \num{+20063.15} & \num{+4581.86} & \num{+10828.79} \\ 
\hline
\end{tabular}
\label{tab:best_scores}
\end{table*}


\subsubsection*{Influence of Forecasting Errors on Revenue}
\label{sec:influence_forecasting_errors}

\begin{figure*}[htbp]
	\centering
	\includegraphics[width=\textwidth]{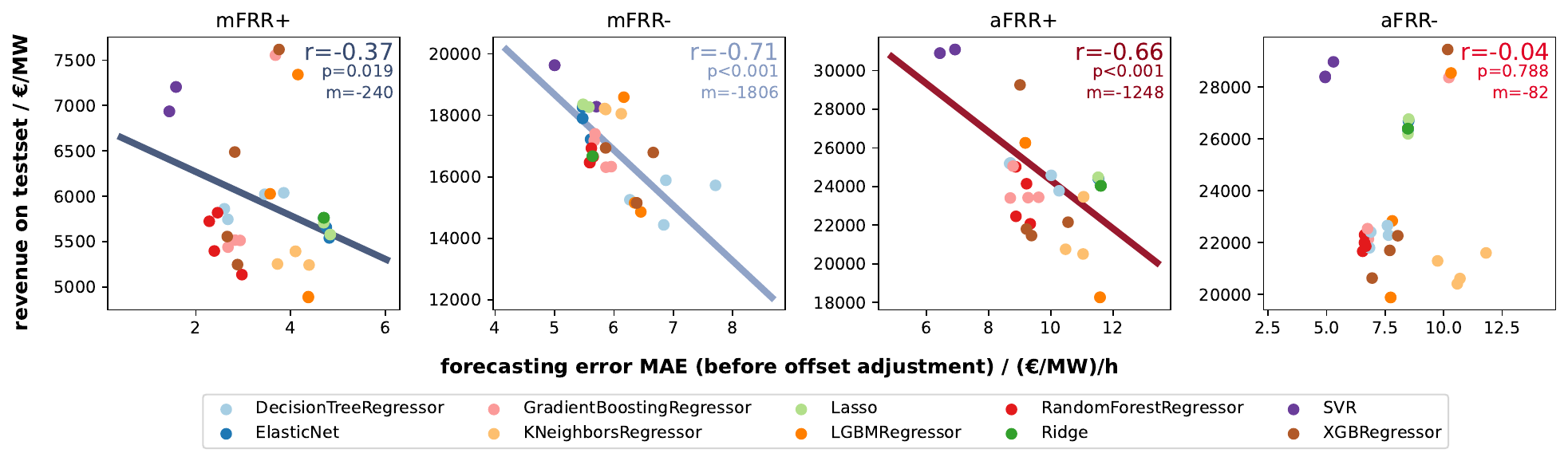}
	\caption{Relation between forecasting errors and final testset output revenue.}
	\label{fig:forecasting_errors}
\end{figure*}

In the previous section, a major difference was seen in the predictability of the markets and the revenues to be generated.
As a final investigation, the influence of the forecasting performance, i.e., the forecasting error, on the revenues to be generated will be examined. 
To this end, the initial model errors (without offset adjustment) were calculated for each of the four markets. This measures of how well the model can learn on the test data and transfer this to the training data. The revenue after offset adjustment was then evaluated for each of these models.
Figure~\ref{fig:forecasting_errors} shows these results for the four markets as a scatter plot. In addition, the Pearson Correlation $r$ and the $p$-value were calculated.

For the \ac{mFRR}-~market, a strong negative correlation ($r_{\texttt{mFFR-}}$=-0.714, $p_{\texttt{mFFR-}}<0.0001$) between the model forecast errors and the generated revenue is measured:
Based on a linear fit, each percent in model accuracy improvement (\ac{MAE}) results in a \SI{1806}{\euroMW} revenue increase on the testset period. 
This equivalents a \num{3631}€/MW yearly revenue increase.
A similar characteristic holds true for the \ac{aFRR}+ market, with a negative correlation ($r_{\texttt{aFFR+}}$=-0.663, $p_{\texttt{aFFR+}}<0.0001$) and a \SI{1248}{\euroMW} testset (respectively  \SI{2509}{\euroMW} yearly) revenue increase for each percent error reduction.
For the \ac{mFRR}+ only limited correlation is measured ($r_{\texttt{mFFR+}}=-0.370$, $p_{\texttt{mFRR+}}=0.019$), having a smaller error-improvement slope of \SI{240}{\euroMW} (\SI{483}{\euroMW} yearly). 
For the \ac{aFRR}- market, no correlation is measured ($r_{\texttt{aFFR-}}=-0.044$, $p_{\texttt{aFRR-}}=0.788$). 

After conducting a comprehensive analysis in the \texttt{aFRR} market and comparing all models across different markets, the authors were unable to identify a single underlying cause for this behavior. 
However, given the limited number of model classes, the observed correlation in all other markets suggests the possibility of a statistical anomaly. 
Expanding the set of models could provide further insights and potentially challenge this hypothesis. 
Moreover, the impact of offset adjustments varies across model architectures due to differences in their residual distributions. 
In the specific case of the latter market, this effect may be the dominant factor.

\section{Further Improvements and Outlook}
\label{sec:Outlook}

%

%

In light of future work and outlook, several major improvements have been identified to further enhance the forecasting performance in the German control reserve market. These go beyond the obvious approaches of expanding parameters for hyperparameter tuning and customized preprocessing, such as removing spikes through data cropping to eliminate bias.

\textbf{First}, the adoption of value-based loss functions is proposed. Since the forecasting target operates within a pay-as-bid auction framework, traditional loss functions that focus solely on minimizing the difference between predicted and actual values may not adequately capture the economic implications. Specifically, bids that exceed the auction value yield no returns, rendering standard loss functions less effective. By integrating this behavior directly into the loss functions, the forecasting methods can better reflect the true economic incentives and constraints of the market.

\textbf{Second}, a systematic shortening of the training dataset is suggested. Currently, models are trained using the entire historical dataset. However, market dynamics may shift over time, making older data less relevant. By focusing on more recent data, such as the last six months, the models can be more responsive to current market conditions, potentially improving forecasting accuracy.
Furthermore, our experiments assume a static market. 
To challenge this assumption, varying the training and evaluation periods could provide valuable insights—for instance, training on data from 2022 to 2024 while evaluating on 2021 data.

\textbf{Third}, it is recommended that additional and more advanced forecasting models be explored. 
Implementing techniques like \ac{LSTM} networks or \acp{RNN} could enhance the models' ability to capture complex patterns in the data. 
Leveraging these sophisticated models increases the probability of identifying a globally optimal forecasting model, thereby improving overall forecasting performance.

\textbf{Fourth}, the incorporation of additional exogenous variables through multivariate forecasting is advocated. By integrating external data sources such as weather information or prices from other energy markets like the day-ahead market, the forecasting models can capture more complex dependencies and patterns that univariate models might miss. Weather conditions significantly impact energy supply and demand, influencing market prices. Similarly, fluctuations in other energy markets can provide insights into potential price movements in the control reserve market. Including these variables as inputs could enhance the models' predictive capabilities and lead to more accurate and robust forecasts.

Furthermore, the proposed method is a general approach for forecasting in pay-as-bid markets. Therefore, in the future, this method could also be applied to other markets that operate under the pay-as-bid mechanism. Additionally, it can be transferred to other ancillary service markets in different countries with similar market structures.


\section{Conclusion}
\label{sec:conclusion}

We developed a methodology for forecasting bidding prices in pay-as-bid ancillary service markets, focusing on the German control reserve market. 
We evaluated our method across four markets: \ac{aFRR}+, \ac{aFRR}--, \ac{mFRR}+, and \ac{mFRR}--.

Our results show that model selection is key to forecasting performance. 
The Support Vector Regressor and the XGBRegressor consistently outperformed other models in accuracy. Hyperparameter tuning and frequent retraining had minimal impact on performance.

Offset adjustment proved to be the most significant factor in improving revenue outcomes. By applying a systematic offset to the predicted prices, we addressed the asymmetry in the revenue function of pay-as-bid markets, where overbidding yields no returns. This adjustment contributed the most to the overall revenue increase. By applying the offset adjustment, we were able to increase revenues by approximately \SI{27.4}{\%} to \SI{37.31}{\%} (meaning a yearly increment between \SI{4581}{\euroMW} and \SI{23326}{\euroMW}, depending on the market, compared to the best benchmark models.

When analyzing the relationship between model forecasting errors without and the resulted revenue after offset adjustment, a positive correlation is detected in the \ac{mFRR}+,  \ac{mFRR}- and \ac{aFRR}+ markets, motivating for improvements of the model forecasting performances.
According to a linear fit, a reduction of 1€/MW MAE model forecasting error statistically leads to a yearly revenue increase between \SI{483}{\euroMW} and \SI{3631}{\euroMW}.  

In summary, selecting an appropriate forecasting model like \ac{SVR}, combined with offset adjustment, is more effective than extensive hyperparameter tuning or frequent retraining for pay-as-bid ancillary service markets. This approach helps industrial participants optimize their bidding strategies and increase their revenues while supporting grid stability.

\subsection*{Acronyms}
\begin{acronym}[Bash]
 \acro{FCR}{Frequency Containment Reserves}
 \acro{mFRR}{manual Frequency Restoration Reserves}
 \acro{aFRR}{automatic Frequency Restoration Reserves}

  \acro{kNNR}{KNeighborsRegressor}
  \acro{MAPE}{Mean Absolute Percentage Error}
  \acro{MAE}{Mean Absolute Error}
  \acro{API}{Application Programming Interface}
  \acro{SVR}{Support Vector Regressor} 
  \acro{SVM}{Support Vector Machine} 
  \acro{XGB}{XGBoostRegressor} 
  \acro{DTR}{DecisionTreeRegressor}
  \acro{LSTM}{Long Short-Term Memory}
  \acro{RNN}{Recurrent Neural Network}
  
\end{acronym}

\subsection*{Declaration of generative AI and AI-assisted technologies in the writing process}
During the preparation of this work the authors used ChatGPT for proofreading and style improvement. 
After using this tool/service, the authors reviewed and edited the content as needed and take full responsibility for the content of the publication.

\section*{CRediT authorship contribution statement}

\textbf{Vincent Bezold:} Data Curation, Methodology, Software, Validation, Visualization, Writing – Original Draft Preparation, Writing – Review \& Editing.  
\textbf{Lukas Baur:} Data Curation, Methodology, Software, Validation, Visualization, Writing – Original Draft Preparation, Writing – Review \& Editing.  
\textbf{Alexander Sauer:} Funding Acquisition, Project Administration, Resources, Writing – Review \& Editing.

\bibliographystyle{unsrt}


\begin{thebibliography}{10}

\bibitem{ortner2019future}
Andr{\'e} Ortner and Gerhard Totschnig.
\newblock The future relevance of electricity balancing markets in europe-a 2030 case study.
\newblock {\em Energy Strategy Reviews}, 24:111--120, 2019.

\bibitem{Blatt1VDI2020}
Verein{ }Deutscher{ }Ingenieure.
\newblock Vdi-richtlinie 5207 energieflexible fabrik. grundlagen - blatt 1. energy-flexible factory - fundamentals.
\newblock {\em Beuth}, July 2020.

\bibitem{tristan2024methodology}
Alejandro Trist{\'a}n-Jim{\'e}nez.
\newblock {\em Methodology for the systematic identification, evaluation, and management of operation-friendly energy flexibility in industrial systems}.
\newblock PhD thesis, University of Stuttgart, 2024.

\bibitem{poplawski2015forecasting}
Tomasz Pop{\l}awski, Grzegorz Dudek, and Jacek {\L}yp.
\newblock Forecasting methods for balancing energy market in poland.
\newblock {\em International Journal of Electrical Power \& Energy Systems}, 65:94--101, 2015.

\bibitem{wang2018ensemble}
Yi~Wang, Qixin Chen, Mingyang Sun, Chongqing Kang, and Qing Xia.
\newblock An ensemble forecasting method for the aggregated load with subprofiles.
\newblock {\em IEEE Transactions on Smart Grid}, 9(4):3906--3908, 2018.

\bibitem{borges2012evaluating}
Cruz~E Borges, Yoseba~K Penya, and Ivan Fernandez.
\newblock Evaluating combined load forecasting in large power systems and smart grids.
\newblock {\em IEEE Transactions on Industrial Informatics}, 9(3):1570--1577, 2012.

\bibitem{nazir2023forecasting}
Amril Nazir, Abdul~Khalique Shaikh, Abdul~Salam Shah, and Ashraf Khalil.
\newblock Forecasting energy consumption demand of customers in smart grid using temporal fusion transformer (tft).
\newblock {\em Results in Engineering}, 17:100888, 2023.

\bibitem{raza2016recent}
Muhammad~Qamar Raza, Mithulananthan Nadarajah, and Chandima Ekanayake.
\newblock On recent advances in pv output power forecast.
\newblock {\em Solar Energy}, 136:125--144, 2016.

\bibitem{rahimi2023comprehensive}
Negar Rahimi, Sejun Park, Wonseok Choi, Byoungryul Oh, Sookyung Kim, Young-ho Cho, Sunghyun Ahn, Chulho Chong, Daewon Kim, Cheong Jin, et~al.
\newblock A comprehensive review on ensemble solar power forecasting algorithms.
\newblock {\em Journal of Electrical Engineering \& Technology}, 18(2):719--733, 2023.

\bibitem{simankov2023review}
Vladimir Simankov, Pavel Buchatskiy, Semen Teploukhov, Stefan Onishchenko, Anatoliy Kazak, and Petr Chetyrbok.
\newblock Review of estimating and predicting models of the wind energy amount.
\newblock {\em Energies}, 16(16):5926, 2023.

\bibitem{beichter2022net}
Maximilian Beichter, Kaleb Phipps, Martha~Maria Frysztacki, Ralf Mikut, Veit Hagenmeyer, and Nicole Ludwig.
\newblock Net load forecasting using different aggregation levels.
\newblock {\em Energy Informatics}, 5(Suppl 1):19, 2022.

\bibitem{markovics2024short}
D{\'a}vid Markovics, Martin~J{\'a}nos Mayer, and Tam{\'a}s Bessenyei.
\newblock Short-term imbalance forecasting with machine learning for proactive mfrr controlling.
\newblock In {\em 2024 9th International Youth Conference on Energy (IYCE)}, pages 1--6. IEEE, 2024.

\bibitem{klaeboe2015benchmarking}
Gro Kl{\ae}boe, Anders~Lund Eriksrud, and Stein-Erik Fleten.
\newblock Benchmarking time series based forecasting models for electricity balancing market prices.
\newblock {\em Energy Systems}, 6:43--61, 2015.

\bibitem{merten2020automatic}
Michael Merten, Fabian Ruecker, Ilka Schoeneberger, and Dirk~Uwe Sauer.
\newblock Automatic frequency restoration reserve market prediction: Methodology and comparison of various approaches.
\newblock {\em Applied Energy}, 268:114978, 2020.

\bibitem{10161755}
Marija Miletić, Tomislav Capuder, Hrvoje Pandžić, and Ivan Pavić.
\newblock Manual frequency restoration reserve markets: What does the data tell us?
\newblock In {\em 2023 19th International Conference on the European Energy Market (EEM)}, pages 1--6, 2023.

\bibitem{bassy2024viability}
{\'A}lvaro Bassy~Navarro.
\newblock Viability and participation model in balancing markets with wind assets in germany.
\newblock Master's thesis, Universidad Pontificia Comillas, 2024.

\bibitem{Bezold2024-tp}
Vincent Bezold, Alexander Mutz, Philipp Pelger, Kerim Torolsan, and Alexander Sauer.
\newblock Profitability analysis of industrial energy flexibility within the german control reserve market, 2024.

\bibitem{narajewski2022probabilistic}
Micha{\l} Narajewski.
\newblock Probabilistic forecasting of german electricity imbalance prices.
\newblock {\em Energies}, 15(14):4976, 2022.

\bibitem{hirth2015balancing}
Lion Hirth and Inka Ziegenhagen.
\newblock Balancing power and variable renewables: Three links.
\newblock {\em Renewable and Sustainable Energy Reviews}, 50:1035--1051, 2015.

\bibitem{regelleistung_historie}
{regelleistung.net}.
\newblock Der regelreservemark. historie: Gemeinsame beschaffung von regelreserve der deutschen Ünb, 2024.
\newblock Accessed: 2024-12-31.

\bibitem{lucas2020price}
Alexandre Lucas, Konstantinos Pegios, Evangelos Kotsakis, and Dan Clarke.
\newblock Price forecasting for the balancing energy market using machine-learning regression.
\newblock {\em Energies}, 13(20):5420, 2020.

\bibitem{vandezande2010well}
Leen Vandezande, Leonardo Meeus, Ronnie Belmans, Marcelo Saguan, and Jean-Michel Glachant.
\newblock Well-functioning balancing markets: A prerequisite for wind power integration.
\newblock {\em Energy policy}, 38(7):3146--3154, 2010.

\bibitem{Diakonikolas2018Efficient}
Ilias Diakonikolas, Weihao Kong, and Alistair Stewart.
\newblock Efficient algorithms and lower bounds for robust linear regression.
\newblock {\em ArXiv}, abs/1806.00040, 2018.

\bibitem{Tay2021Elastic}
J.~K. Tay, B.~Narasimhan, and T.~Hastie.
\newblock Elastic net regularization paths for all generalized linear models.
\newblock {\em Journal of statistical software}, 106, 2021.

\bibitem{James2021Moving}
Gareth~M. James, D.~Witten, T.~Hastie, and R.~Tibshirani.
\newblock Moving beyond linearity.
\newblock {\em Springer Texts in Statistics}, 2021.

\bibitem{Andersen2009Nonparametric}
Robert Andersen.
\newblock Nonparametric methods for modeling nonlinearity in regression analysis.
\newblock {\em Review of Sociology}, 35:67--85, 2009.

\bibitem{Zeng2019Multiple}
Shan Zeng, C.~Gao, Xiuying Wang, Liang Jiang, and D.~Feng.
\newblock Multiple kernel-based discriminant analysis via support vectors for dimension reduction.
\newblock {\em IEEE Access}, 7:35418--35430, 2019.

\bibitem{Nguyen2022Robust}
Khanh-Duy Nguyen, Trung Le, T.~Nguyen, Geoffrey~I. Webb, and Dinh~Q. Phung.
\newblock Robust variational learning for multiclass kernel models with stein refinement.
\newblock {\em IEEE Transactions on Knowledge and Data Engineering}, 34:4425--4438, 2022.

\bibitem{Bentejac2019A}
Candice Bentéjac, Anna Csörgo, and Gonzalo Martínez-Muñoz.
\newblock A comparative analysis of gradient boosting algorithms.
\newblock {\em Artificial Intelligence Review}, 54:1937 -- 1967, 2019.

\bibitem{Neyshabur2017Exploring}
Behnam Neyshabur, Srinadh Bhojanapalli, D.~McAllester, and N.~Srebro.
\newblock Exploring generalization in deep learning.
\newblock pages 5947--5956, 2017.

\bibitem{Whang2020Data}
Steven~Euijong Whang and Jae-Gil Lee.
\newblock Data collection and quality challenges for deep learning.
\newblock {\em Proceedings of the VLDB Endowment}, 13:3429 -- 3432, 2020.

\bibitem{Shen2024On}
Li~Shen, Yan Sun, Zhiyuan Yu, Liang Ding, Xinmei Tian, and Dacheng Tao.
\newblock On efficient training of large-scale deep learning models.
\newblock {\em ACM Computing Surveys}, 2024.

\bibitem{nextkraftwerke_regelenergie}
{Next Kraftwerke}.
\newblock Aktivierung der verschiedenen regelenergiearten.
\newblock 2025.
\newblock Accessed: 2025-01-03.

\bibitem{regelleistung_FCR}
{regelleistung.net}.
\newblock Frequency containment reserve, 2024.
\newblock Accessed: 2024-12-31.

\bibitem{regelleistung_aFRR}
{regelleistung.net}.
\newblock automatic frequency restoration reserve, 2024.
\newblock Accessed: 2024-12-31.

\bibitem{regelleistung_mFRR}
{regelleistung.net}.
\newblock manual frequency restoration reserve, 2024.
\newblock Accessed: 2024-12-31.

\bibitem{regelleistungBeschaffung}
{regelleistung.net}.
\newblock {B}eschaffung {R}egelleistung \& {R}egelarbeit, 2024.
\newblock Accessed: 2024-10-22.

\bibitem{regelleistungStartseite}
{regelleistung.net}.
\newblock Regelleistung.net {S}tartseite, 2024.
\newblock Accessed: 2024-10-22.

\bibitem{ertz2022}
TransnetBW Amperion, Tennet.
\newblock Präqualifikationsverfahren für regelreserveanbieter: Pq-anforderungen, 2022.
\newblock Ertz.

\bibitem{regelleistung_Download}
{regelleistung.net}.
\newblock Regelleistung.net – data center, 2024.
\newblock Accessed: 2024-10-22.

\bibitem{scikit-learn}
F.~Pedregosa, G.~Varoquaux, A.~Gramfort, V.~Michel, B.~Thirion, O.~Grisel, M.~Blondel, P.~Prettenhofer, R.~Weiss, V.~Dubourg, J.~Vanderplas, A.~Passos, D.~Cournapeau, M.~Brucher, M.~Perrot, and E.~Duchesnay.
\newblock Scikit-learn: Machine learning in {P}ython.
\newblock {\em Journal of Machine Learning Research}, 12:2825--2830, 2011.

\bibitem{toner2024analysislineartimeseries}
William Toner and Luke Darlow.
\newblock An analysis of linear time series forecasting models, 2024.

\bibitem{green2015simple}
Kesten~C Green and J~Scott Armstrong.
\newblock Simple versus complex forecasting: The evidence.
\newblock {\em Journal of Business Research}, 68(8):1678--1685, 2015.

\bibitem{nti2020electricity}
Isaac~Kofi Nti, Moses Teimeh, Owusu Nyarko-Boateng, and Adebayo~Felix Adekoya.
\newblock Electricity load forecasting: a systematic review.
\newblock {\em Journal of Electrical Systems and Information Technology}, 7:1--19, 2020.

\bibitem{nti2021performance}
Isaac~Kofi Nti, Owusu Nyarko-Boateng, Justice Aning, et~al.
\newblock Performance of machine learning algorithms with different k values in k-fold crossvalidation.
\newblock {\em International Journal of Information Technology and Computer Science}, 13(6):61--71, 2021.

\end{thebibliography}



\newpage

\appendix
\section{Used Grid Search Parameters}
\label{sec:used_grid_search_parameters}

A listing of the hyperparameter for parameter optimization is given in Table~\ref{tab:param_grid}.
\begin{table*}[htbp]
\centering
\caption{Used parameters for Grid Search.}
\begin{tabular}{|l|l|l|}
\hline
\textbf{Model}                             & \textbf{Hyperparametabstrer} & \textbf{Search Space}   \\ \hline
\multirow{2}{*}{SVR}                       & C                        & {[}0.1, 1, 10{]}        \\ \cline{2-3} 
                                           & epsilon                  & {[}0.1, 0.2, 0.5{]}     \\ \hline
\multirow{3}{*}{GradientBoostingRegressor} & n\_estimators            & {[}100, 200{]}          \\ \cline{2-3} 
                                           & learning\_rate           & {[}0.01, 0.1{]}         \\ \cline{2-3} 
                                           & max\_depth               & {[}3, 5{]}              \\ \hline
\multirow{3}{*}{XGBRegressor}              & n\_estimators            & {[}100, 200{]}          \\ \cline{2-3} 
                                           & learning\_rate           & {[}0.01, 0.1{]}         \\ \cline{2-3} 
                                           & max\_depth               & {[}3, 5{]}              \\ \hline
\multirow{3}{*}{LGBMRegressor}             & n\_estimators            & {[}100, 200{]}          \\ \cline{2-3} 
                                           & learning\_rate           & {[}0.01, 0.1{]}         \\ \cline{2-3} 
                                           & num\_leaves              & {[}31, 50{]}            \\ \hline
\multirow{2}{*}{RandomForestRegressor}     & n\_estimators            & {[}100, 200{]}          \\ \cline{2-3} 
                                           & max\_depth               & {[}None, 10, 20{]}      \\ \hline
\multirow{2}{*}{DecisionTreeRegressor}     & max\_depth               & {[}None, 10, 20{]}      \\ \cline{2-3} 
                                           & min\_samples\_split      & {[}2, 10{]}             \\ \hline
\multirow{2}{*}{KNeighborsRegressor}       & n\_neighbors             & {[}3, 5, 7{]}           \\ \cline{2-3} 
                                           & weights                  & {[}uniform, distance{]} \\ \hline
Ridge                                      & alpha                    & {[}0.1, 1.0, 10.0{]}    \\ \hline
Lasso                                      & alpha                    & {[}0.1, 1.0, 10.0{]}    \\ \hline
\multirow{2}{*}{ElasticNet}                & alpha                    & {[}0.1, 1.0, 10.0{]}    \\ \cline{2-3} 
                                           & l1\_ratio                & {[}0.1, 0.5, 0.9{]}     \\ \hline
\end{tabular}
\label{tab:param_grid}
\end{table*}

\section{Hyperparameter Tuning: Performance Comparisons}
\label{sec:appendix_hyperparam_images}
\begin{figure*}[htbp]
	\centering
	\includegraphics[width=\textwidth]{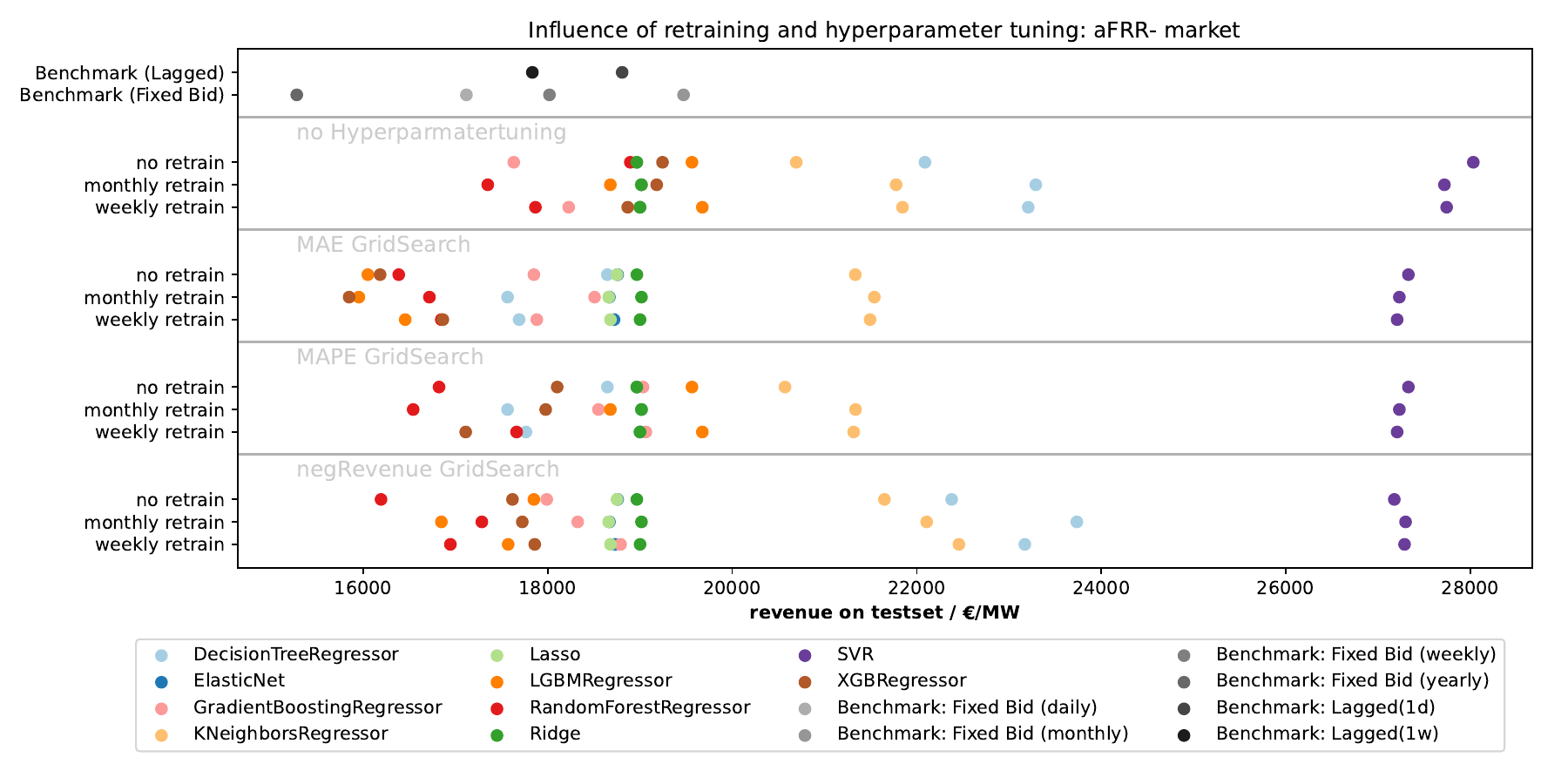}
	\caption{Performance of the models with different hyperparameter tuning loss functions compared with the benchmarks (results for \ac{aFRR}-\ market).}
	\label{fig:appendix_models_hyperparam2}
\end{figure*}

\begin{figure*}[htbp]
	\centering
	\includegraphics[width=\textwidth]{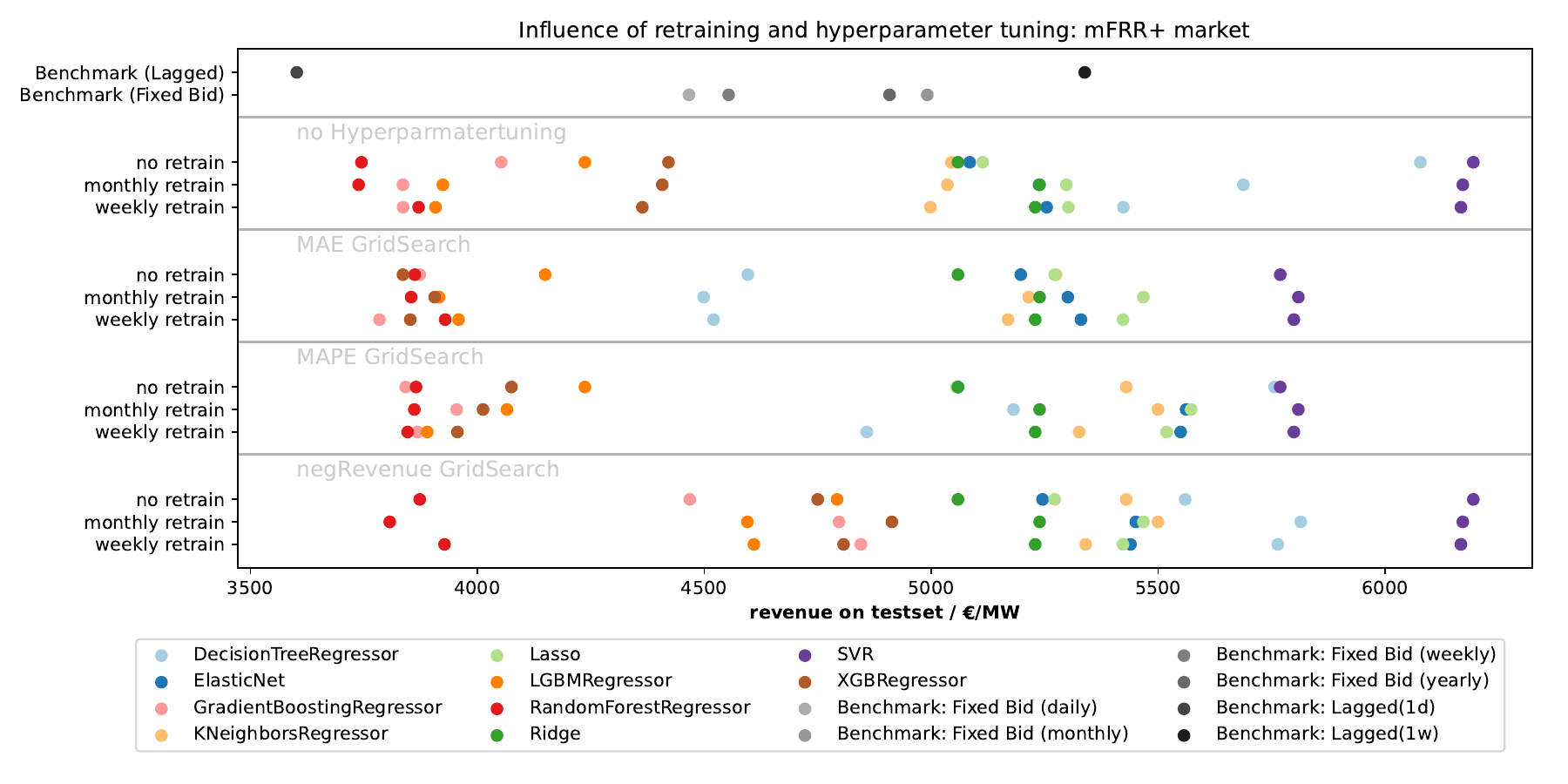}
	\caption{Performance of the models with different hyperparameter tuning loss functions compared with the benchmarks (results for \ac{mFRR}+\ market).}
	\label{fig:appendix_models_hyperparam3}
\end{figure*}

\begin{figure*}[htbp]
	\centering
	\includegraphics[width=\textwidth]{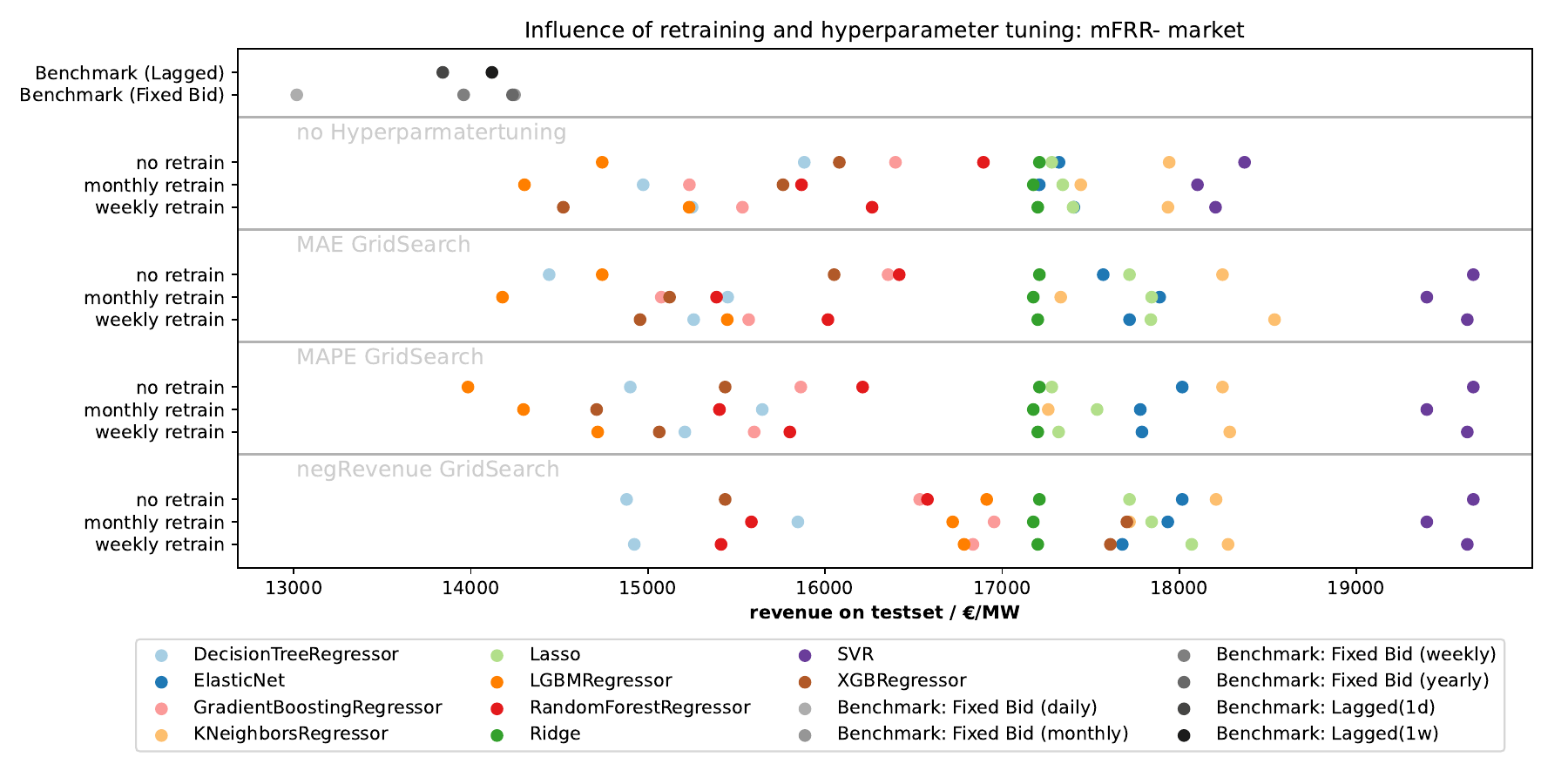}
	\caption{Performance of the models with different hyperparameter tuning loss functions compared with the benchmarks (results for \ac{mFRR}-\ market).}
	\label{fig:appendix_models_hyperparam4}
\end{figure*}

The influence of the different hyperparameter tuning losses on the model performances in the other markets are shown in Figures~\ref{fig:appendix_models_hyperparam2},\ref{fig:appendix_models_hyperparam3}, and \ref{fig:appendix_models_hyperparam4}. 


\section{Offset Choice: Examples for SVM on other Markets}
\begin{figure*}[htbp]
	\centering
	\includegraphics[width=0.7\linewidth]{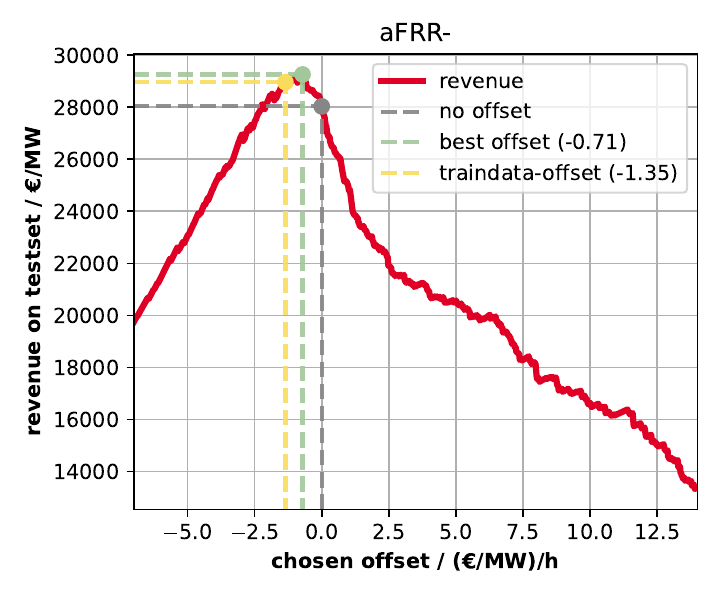}
	\caption{Visualization of the total test data revenue by applying a forecast offset $\delta$ in the \ac{aFRR}- market using the \ac{SVR} model.}
	\label{fig:appendix_offset_adjust__offset_over_revenue1}
\end{figure*}

\begin{figure*}[htbp]
	\centering
	\includegraphics[width=0.7\linewidth]{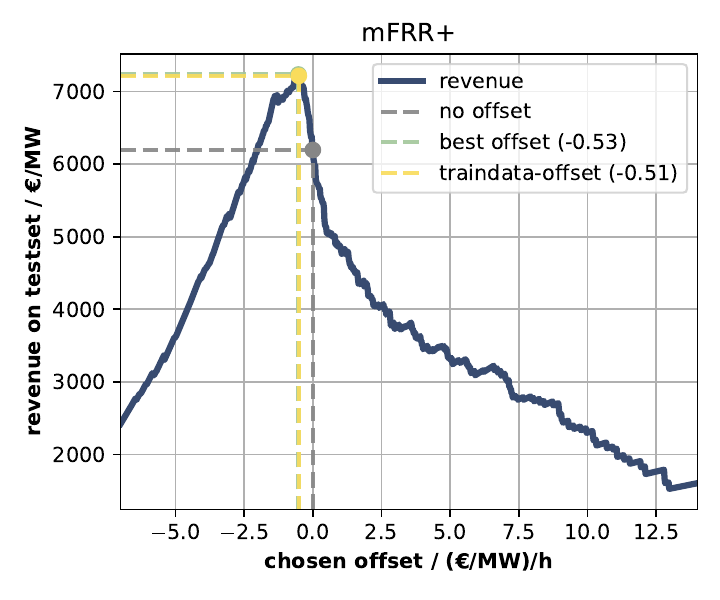}
	\caption{Visualization of the total test data revenue by applying a forecast offset $\delta$ in the \ac{mFRR}+ market using the \ac{SVR} model.}
	\label{fig:appendix_offset_adjust__offset_over_revenue2}
\end{figure*}

\begin{figure*}[htbp]
	\centering
	\includegraphics[width=0.7\linewidth]{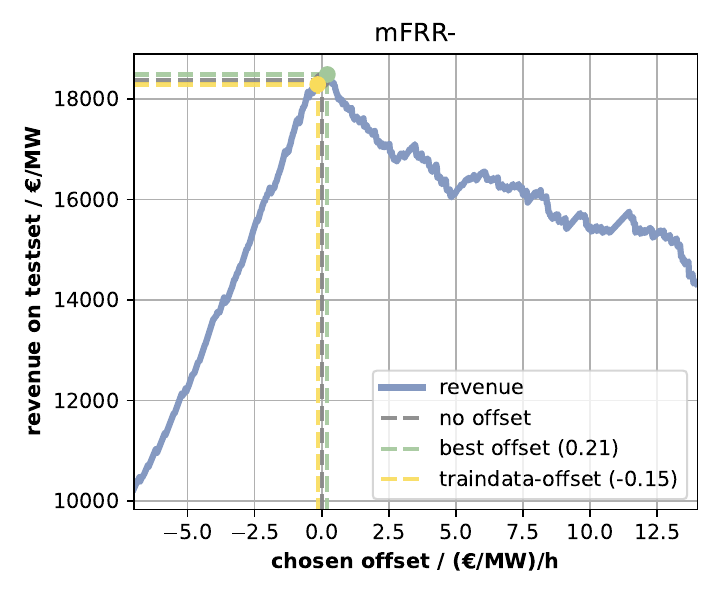}
	\caption{Visualization of the total test data revenue by applying a forecast offset $\delta$ in the \ac{mFRR}- market using the \ac{SVR} model.}
	\label{fig:appendix_offset_adjust__offset_over_revenue3}
\end{figure*}

The offset choise selection based on the training set data using the \ac{SVR} model in the other markets is shown in Figures~\ref{fig:appendix_offset_adjust__offset_over_revenue1},\ref{fig:appendix_offset_adjust__offset_over_revenue2}, and \ref{fig:appendix_offset_adjust__offset_over_revenue3}. 

\section{Offset Adjustment: Influence on Model Performances in other Markets}
\label{sec:appendix_offset_adjustment_performance_change}
\begin{figure*}[htbp]
	\centering
	\includegraphics[width=\textwidth]{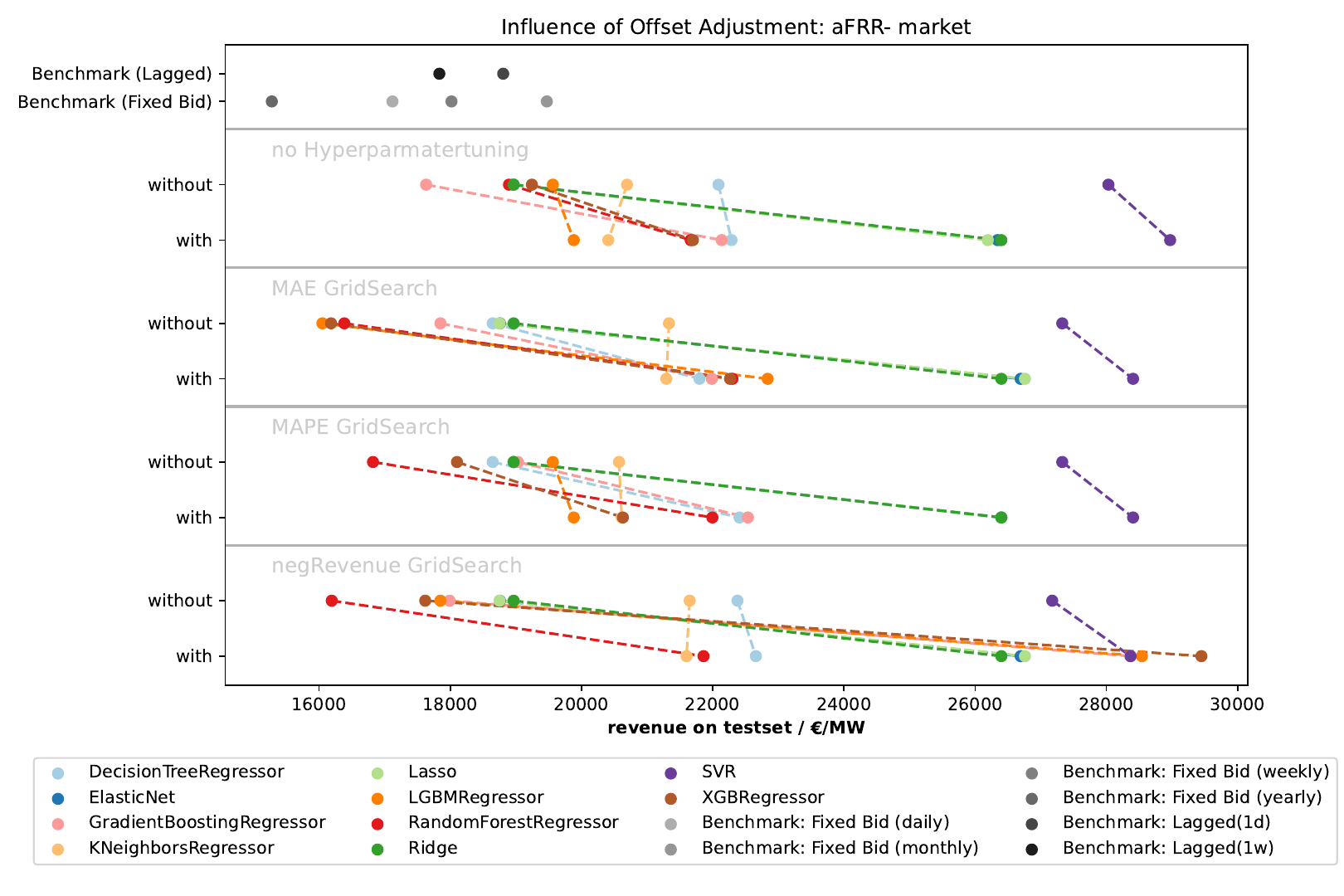}
	\caption{Performance comparison of the models with and without applying offset adjustment (results for \ac{aFRR}-\ market).}
	\label{fig:appendix_with_and_without_offset_adjustment2}
\end{figure*}

\begin{figure*}[htbp]
	\centering
	\includegraphics[width=\textwidth]{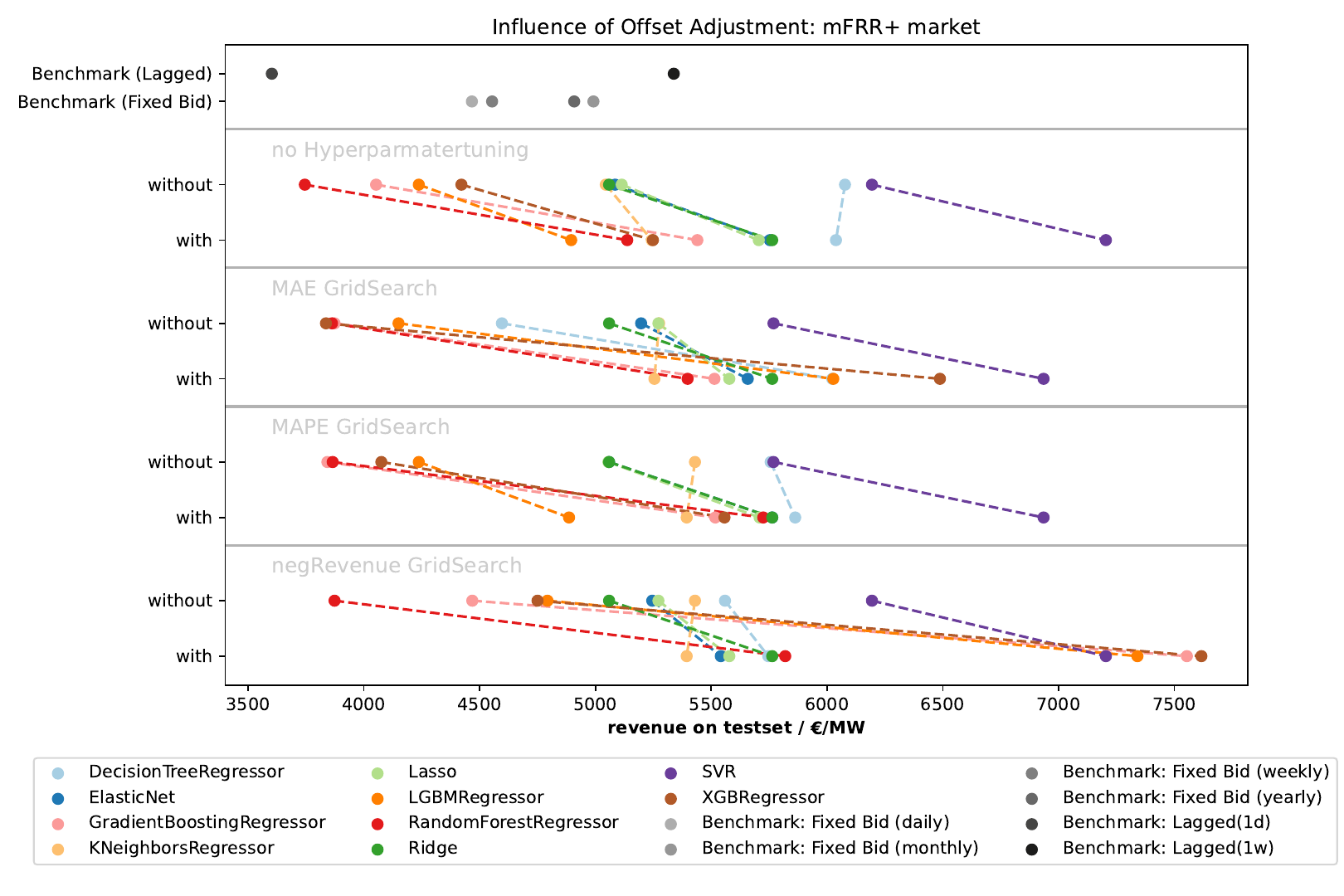}
	\caption{Performance comparison of the models with and without applying offset adjustment (results for \ac{mFRR}+\ market).}
	\label{fig:appendix_with_and_without_offset_adjustment3}
\end{figure*}

\begin{figure*}[htbp]
	\centering
	\includegraphics[width=\textwidth]{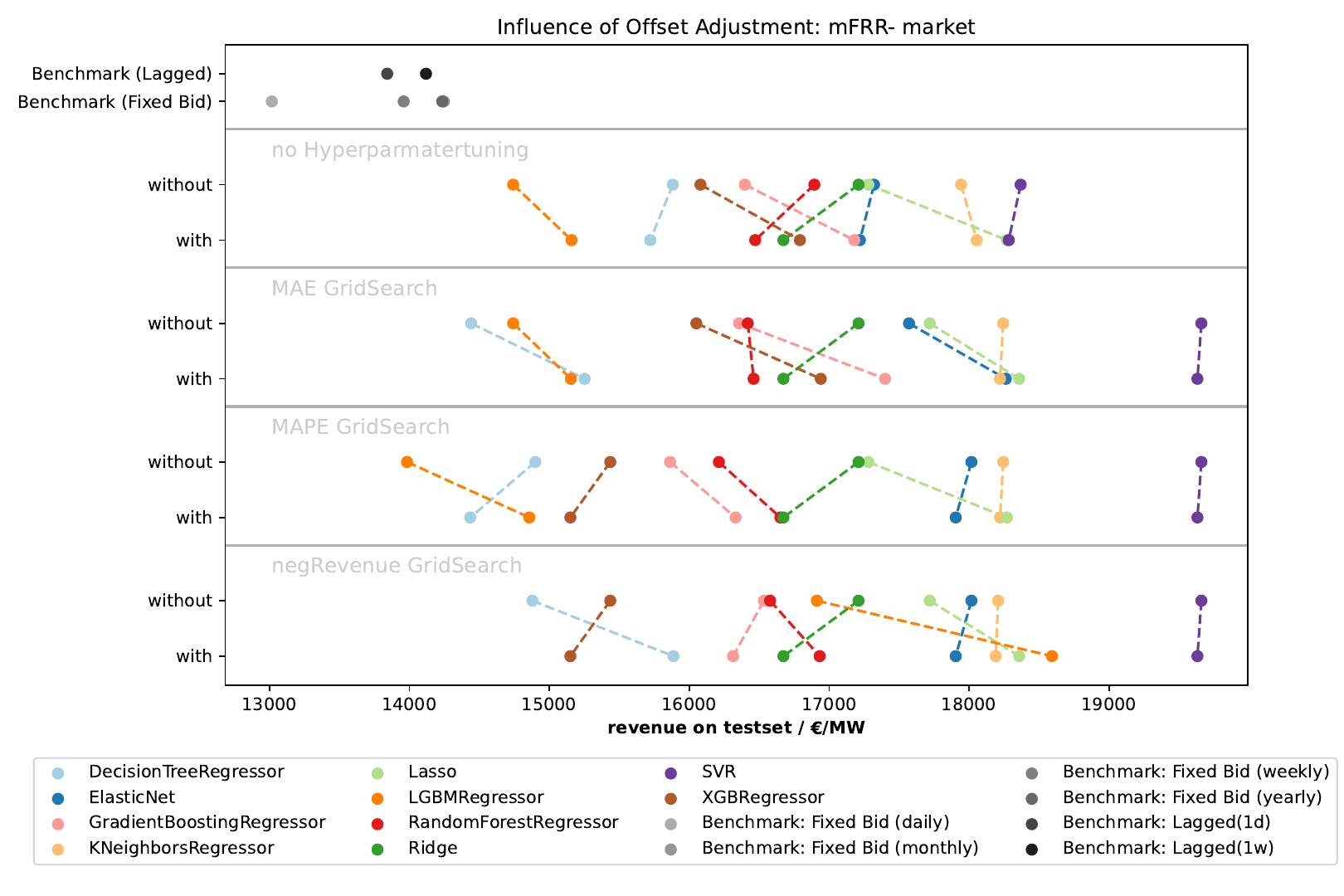}
	\caption{Performance comparison of the models with and without applying offset adjustment (results for \ac{mFRR}-\ market).}
	\label{fig:appendix_with_and_without_offset_adjustment4}
\end{figure*}

The influence of the offset adjustment on the models in the other markets are shown in Figures~\ref{fig:appendix_with_and_without_offset_adjustment2},\ref{fig:appendix_with_and_without_offset_adjustment3}, and \ref{fig:appendix_with_and_without_offset_adjustment4}. 


\end{document}